\documentclass[10pt,twocolumn,letterpaper]{article}

\usepackage[pagenumbers]{cvpr} %
\definecolor{cvprblue}{rgb}{0.21,0.49,0.74}
\usepackage[pagebackref,breaklinks,colorlinks,allcolors=cvprblue]{hyperref}

\usepackage{makecell}

\title{StyleMaster: Stylize Your Video with Artistic Generation and Translation}

\author{Zixuan Ye\textsuperscript{1}\textsuperscript{\dag} \quad Huijuan Huang\textsuperscript{2}\textsuperscript{*} \quad Xintao Wang\textsuperscript{2} \quad Pengfei Wan\textsuperscript{2} \quad Di Zhang\textsuperscript{2} \quad Wenhan Luo\textsuperscript{1}\textsuperscript{*}\\
\textsuperscript{1} Hong Kong University of Science and Technology \quad \textsuperscript{2} KuaiShou Technology
}

\begin{document}

\twocolumn[{%
\renewcommand\twocolumn[1][]{#1}%
\maketitle
\begin{center}
    \centering
    \captionsetup{type=figure}
    \includegraphics[width=1\linewidth]{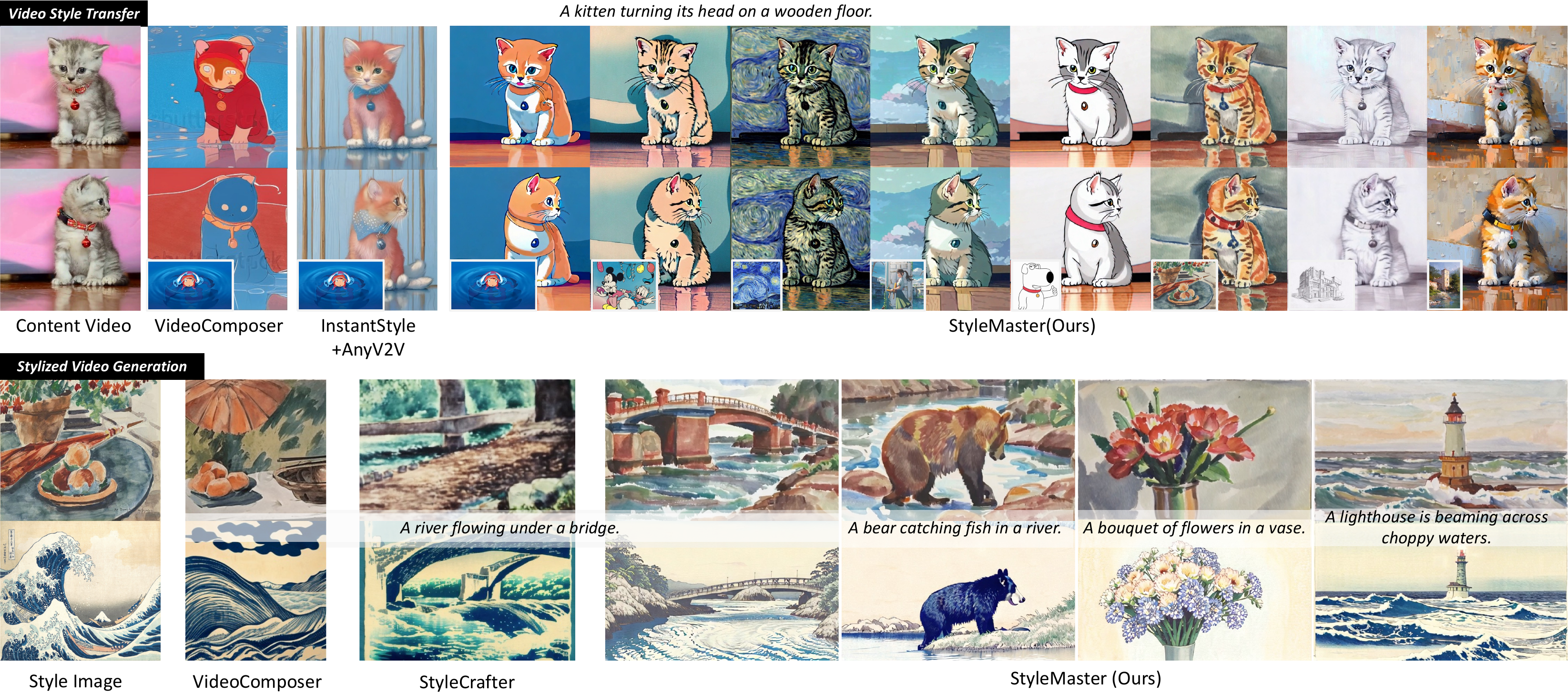}
    \captionof{figure}{\textbf{Our StyleMaster demonstrates superior video style transfer and stylized generation}. The top section shows our method effectively applying various styles to videos, outperforming VideoComposer~\cite{wang2024videocomposer} and the combination of InstantStyle~\cite{instantstyle} with AnyV2V~\cite{anyv2v}. The bottom highlights our high-quality text-driven stylized synthesis, surpassing VideoComposer~\cite{wang2024videocomposer} and StyleCrafter~\cite{stylecrafter}.}
\end{center}%
}]

\renewcommand{\thefootnote}{\fnsymbol{footnote}}
\setcounter{footnote}{0}
\footnotetext[1]{Corresponding authors.}
\footnotetext[2]{Work done during internship at KwaiVGI, Kuaishou Technology.}

\begin{abstract}

Style control has been popular in video generation models. Existing methods often generate videos far from the given style, cause content leakage, and struggle to transfer one video to the desired style. Our first observation is that the style extraction stage matters, whereas existing methods emphasize global style but ignore local textures. In order to bring texture features while preventing content leakage, we filter content-related patches while retaining style ones based on prompt-patch similarity; for global style extraction, we generate a paired style dataset through model illusion to facilitate contrastive learning, which greatly enhances the absolute style consistency. Moreover, to fill in the image-to-video gap, we train a lightweight motion adapter on still videos, which implicitly enhances stylization extent, and enables our image-trained model to be seamlessly applied to videos.
Benefited from these efforts, our approach, StyleMaster, not only achieves significant improvement in both style resemblance and temporal coherence, but also can easily generalize to video style transfer with a gray tile ControlNet. Extensive experiments and visualizations demonstrate that StyleMaster significantly outperforms competitors, effectively generating high-quality stylized videos that align with textual content and closely resemble the style of reference images. Our project page is at \href{https://zixuan-ye.github.io/stylemaster}{https://zixuan-ye.github.io/stylemaster}.

\end{abstract}

\section{Introduction}
\label{sec:intro}

\begin{figure}
    \centering
    \includegraphics[width=1\linewidth]{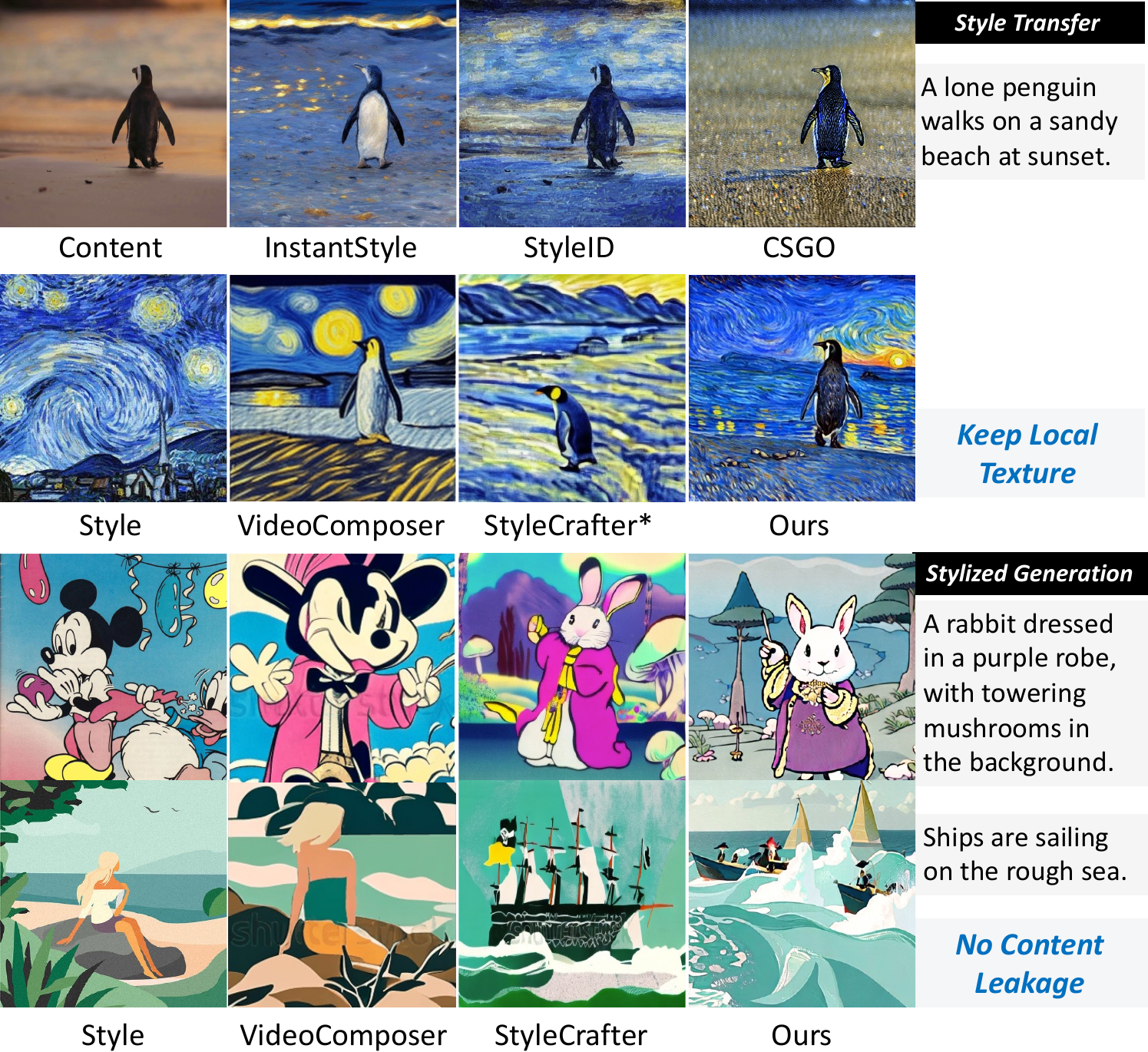}
    \caption{\textbf{Existing image and video stylization methods either fail in keeping local texture or suffer from content leakage.} Note: * means StyleCrafter does not support transfer, we use text and reference style image to generate results.}
    \label{fig:texture}
\end{figure}

Video generation~\cite{sora,videocrafter1,cogvideo,latte} has witnessed great success promoted by diffusion models~\cite{stablediffusion,dit,song2020score,lipman2022flow,ma2024sit,ho2020denoising}, which also bring about great controllability. Wherein, the style control, \textit{i.e.}, to generate or translate a video to the same style as a given reference image, is of great interest and importance but less developed. Several instances exhibited in Fig.~\ref{fig:texture} reveals that current methods often struggle to preserve the local textures, such as the brush strokes of a Van Gogh painting. Moreover, they fail to properly decouple content and style: they either focus too much on global style while losing texture details, or overuse reference features, leading to excessive copying and content leakage.

We argue that the failure comes largely from inappropriate use of global and texture features, and formulate separate remedies for both. First, to preserve the texture features, we turn to local patches for style guidance. However, directly using all image patch features from CLIP~\cite{clip} can lead to content leakage. Thus, we keep patches of less similarity with text prompts, while discard the rest ones. We empirically find that the selected patches can carry sufficient texture information, without bringing any associated contents. Though the texture features are helpful for style representations, they are not sufficient to completely represent the style of an image without global information. One naive practice to incorporate global representation can be the global embedding from the CLIP encoder. However, it may easily invite content leakage, like that in VideoComposer~\cite{wang2024videocomposer}. One possible solution for decoupling is a contrastive learning strategy, with samples of the same style as positive, and others as negative~\cite{styletokenizer}. However, existing style datasets cannot even guarantee style consistency within the group, which is sub-optimal to train a style extractor. Inspired by the illusion property demonstrated in VisualAnagrams~\cite{geng2024illusion}, we can generate paired images where one image is a pixel-rearranged version of the other. Therefore, the content in the two images differs, while they share the same style. Unlike other manually collected and grouped datasets, we can generate a dataset of an almost infinite number of such pairs with minimal effort, while ensuring absolute style consistency within the pair. With these data pairs, we can train a strong module to extract the global-style-oriented features. In our practice, instead of fine-tuning the CLIP model, we opt to train a projection module after CLIP to ensure the generalization ability. 

With the global and local features, the style information is then injected into the model in an adapter-based mechanism through the dual cross-attention strategy~\cite{ipadapter}. Since the image-only training will cause degradation in motion dynamics of videos, we adopt a motion adapter trained with still videos, inspired by StillMoving~\cite{stillmoving}. During inference, by turning the motion adapter's ratio to negative, the motion quality is enhanced. More importantly, if the videos used for training are all in the real-world domain, the negative ratio implicitly helps to enhance the style extent by leading the generated results away from the real-world domain. 

Beyond stylized generation, we explore video translation as a broader application. While existing methods focus only on stylized video generation or rely on depth-based ControlNet for content control~\cite{stylecrafter,wang2024videocomposer}, we propose a simpler solution: we design a gray tile ControlNet as more accessible yet precise content guidance for video style transfer. Extensive experiments show that, StyleMaster can generates high-quality videos with high style similarity to the reference image and achieve ideal translation results, significantly outperforming other competitors in several stylization tasks. Rich ablation studies are conducted to validate the effectiveness of the proposed modules. In conclusion, our contributions are threefold:

\begin{itemize}
    \item We propose a novel style extraction module with local patch selection to overcome the content leakage in style transfer, and global projection to extract strong style cues.
    \item We are the first to propose using model illusion to generate datasets of paired images with absolute style consistency at nearly no cost. This not only produces accurate style-content decoupling in our approach but also benefits style-related research in the community.
    \item With an adopted motion adapter and gray tile ControlNet, our developed StyleMaster is capable of generating content accurately representing the given reference style in both video generation and video/image style transfer tasks, and more importantly, outperforms other methods significantly as demonstrated by the experimental results. 
\end{itemize}

\section{Related Work}
\label{sec:related}

\subsection{Image Stylization}
The success of generative models~\cite{stablediffusion,videocrafter1,cogvideo} has inspired customized generative models specifying object~\cite{dreambooth,anydoor}, edge~\cite{t2iadapter,controlnet,canny}, layout~\cite{boxdiff,layout}, ID~\cite{li2024photomaker}. Controlling generation with a specific style from a reference image has also garnered significant attention, with numerous studies exploring how to extract the style description from the reference image and how to inject the style cues~\cite{artadapter,styleadapter,qi2024deadiff,caiclap}. Inspired by Textural Inversion (TI)~\cite{ti}, some methods~\cite{styledrop, inst, styleid} optimize a specific textual embedding to represent style. Instead of relying on inversion, IP-Adapter~\cite{ipadapter} trains an image adapter to adapt T2I models to image conditions. However, it cannot decouple the style and content in the reference image, resulting in severe content leakage. 

To address this, InstantStyle~\cite{instantstyle,instantstyleplus} identifies the layer truly impacting stylization and injects style only into the identified layer. However, due to the sub-optimal style extractor, it suffers from poor style precision. StyleTokenizer~\cite{styletokenizer} fine-tunes an image encoder with a manually collected and grouped style dataset, Style30K, in a contrastive training manner. However, Style30K cannot maintain style consistency, which adversely affects the extractor. CSGO~\cite{csgo} creates a triplet dataset consisting of content-style-sample pairs generated by B-LoRA~\cite{blora}, achieving impressive performance. However, it can only extract global representations and fails to preserve local textures. Therefore, we propose to consider both local and global information and create a dataset with absolute style consistency, \textit{i.e.}, performing pixel rearrangement to form a new image, leveraging the model's illusion property~\cite{geng2024illusion}.

\subsection{Video Stylization}
One can achieve video style transfer by applying a frame-by-frame process using an image stylization model. However, this can lead to temporal inconsistency. To address this, early deep learning methods~\cite{chen2017coherent, deng2021arbitrary, gao2020fast,huang2017real} employ optical flow constraints. Generative models have elevated this task to a new level. Given the first and/or the last frame, some Image-to-Video (I2V) methods can create stylized videos~\cite{tooncrafter}. However, under this setting, users cannot specify the style with a reference style image. Some video editing methods, like AnyV2V~\cite{anyv2v}, also attempt to stylize the video given an edited first frame, but it requires an image stylization model to obtain the stylized first frame. In contrast, AnimateDiff~\cite{animatediff} extends the Text-to-Image model into a Text-to-Video model by adding a temporal module. StillMoving~\cite{stillmoving} further free the requirement for video data by training a motion adapter with still videos, enabling easy cooperation with any image customization models~\cite{dreambooth}. %

Some works based on T2V models~\cite{animatediff,videocrafter1} focus on controllable video generation. For example, VideoComposer~\cite{wang2024videocomposer} achieves multiple controls including style control. However, directly injecting all reference image tokens during training causes serious content leakage. Instead, StyleCrafter~\cite{stylecrafter} adopts Q-Former to extract the style descriptions from an image. However, it ignores the local texture, resulting in sub-optimal stylization. Additionally, it focuses on stylized generation only, rather than style transfer, which is an important aspect of video stylization.

\section{Method}

\begin{figure}
    \centering
    \includegraphics[width=1\linewidth]{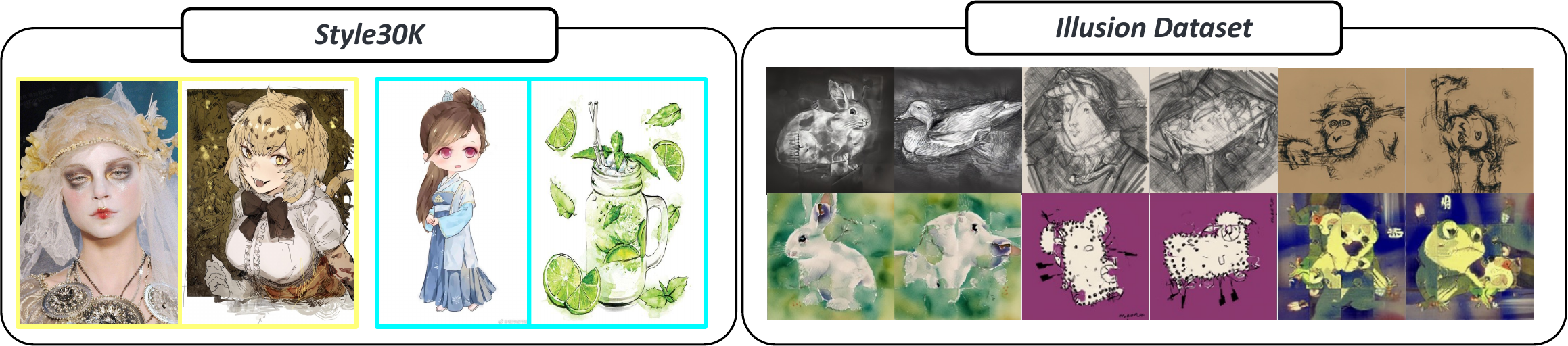}
    \caption{\textbf{Comparison between Style30K with our dataset generated by model illusion}. Style30K cannot ensure consistency within a style group (highlighted by the same color), while ours owns absolute consistency.}
    \label{fig:dataset}
\end{figure}
\label{sec:method}

In this section, we illustrate the components of our StyleMaster. We first construct a contrastive dataset (Sec.~\ref{ssec:construction}) with absolute style consistency, and develop global and local style extraction methods (Sec.~\ref{ssec:global} and Sec.~\ref{ssec:local}) for accurate style representation. 
We mitigate dynamic degradation in Sec.~\ref{ssec:motion}, and introduce a content control mechanism using gray tile guidance in Sec.~\ref{ssec:gray}.

\begin{figure*}
    \centering
    \includegraphics[width=1\linewidth]{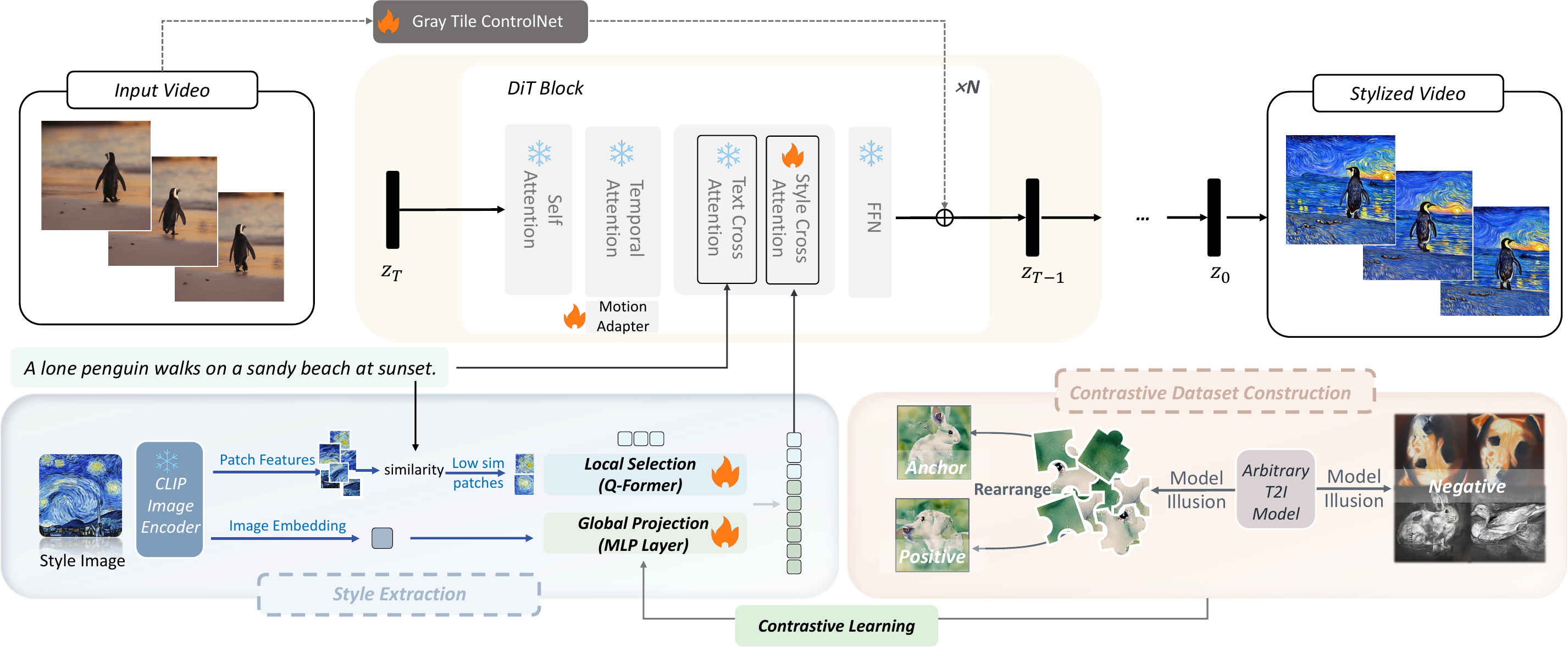}
    \caption{\textbf{The pipeline of our proposed StyleMaster.} We first obtain patch features and image embedding of the style image from CLIP, then we select the patches sharing less similarity with text prompt as texture guidance, and use a global projection module to transform it into global style descriptions. The global projection module is trained with a contrastive dataset constructed by \textit{model illusion} through contrastive learning. The style information is then injected into the model through the decoupled cross-attention. The motion adapter and gray tile ControlNet are used to enhance dynamic quality and enable content control respectively.}
    \label{fig:pipeline}
\end{figure*}

\subsection{Contrastive Dataset Construction}
\label{ssec:construction}
StyleTokenizer~\cite{styletokenizer} collects a style dataset consisting of $30$K style images, groups them into about $30$ style groups, and uses this dataset to finetune the CLIP through contrastive learning. In this step, the quality of the style dataset is crucial, as it largely determines the final capability of the extractor. However, as shown in Fig.~\ref{fig:dataset}, the style consistency within a group cannot be guaranteed. Specifically, the first two images highlighted by yellow bounding boxes illustrate this inconsistency: one belongs to the real-world domain, while the other is from the animation domain, yet both are classified as the same style.  Moreover, the process of collecting and grouping is labor-intensive. Therefore, a more efficient method to obtain style data is required.

We draw inspiration from the success of model illusion~\cite{geng2024illusion}, which uses pretrained T2I models to create optical illusions. To be specific, given an arbitrary T2I model, during the sampling process, we copy and change the view (\textit{e.g.}, rotation, flip) of noisy image to form a parallel process, then use two different prompts to guide the noise prediction of the two noisy images. Then we change the predicted noise back to its original view and add the two predicted noises to form the output noise. In this way, the generated images can change appearance when pixels are rearranged. Based on this, to generate the dataset, we create two lists: one containing objects and the other containing style descriptions. We then randomly select a style and two objects to generate paired images, \textit{e.g}, as shown in Fig.~\ref{fig:pipeline}, the prompts are ``a watercolor painting of a dog" and ``a watercolor painting of a rabbit". Since the paired images in model illusion are merely pixel rearrangements, we can ensure style consistency within a group. Leveraging this property, we can automatically generate an infinite amount of data with no effort.

\subsection{Extract Global Description}
\label{ssec:global}

Rather than fine-tuning the entire CLIP image encoder like StyleTokenizer~\cite{styletokenizer}, which might compromise its generalization ability, we opt for a post-processing module to the image embedding output from CLIP, \textit{i.e.},
\begin{equation}
F_i = \text{CLIP}(I).\texttt{image\_embed}\,,\nonumber
\end{equation}
where $I$ represents the style image, and $F_i\in\mathbb{R}^{1 \times 1024}$. We use a simple MLP layer $f(x) = \text{MLP}(x)$ as a projection to transform the image embedding, which contains both content and style information, into only global style representation. During training, we employ a triplet loss where we treat one image from a paired set as the anchor, its counterpart as the positive sample, and any image outside this pair as the negative sample:
\begin{equation}
\mathcal{L}= \sum_{n=1}^{N} \left[ \| f(F_{i,n}^{\text{anc}}) - f(F_{i,n}^{\text{pos}})\| - \| f(F_{i,n}^{\text{anc}}) - f(F_{i,n}^{\text{neg}}) \| + \alpha \right], \nonumber
\end{equation}
and the margin $\alpha$ defines the distance between samples of different groups. 

\begin{figure}
    \centering
    \includegraphics[width=1\linewidth]{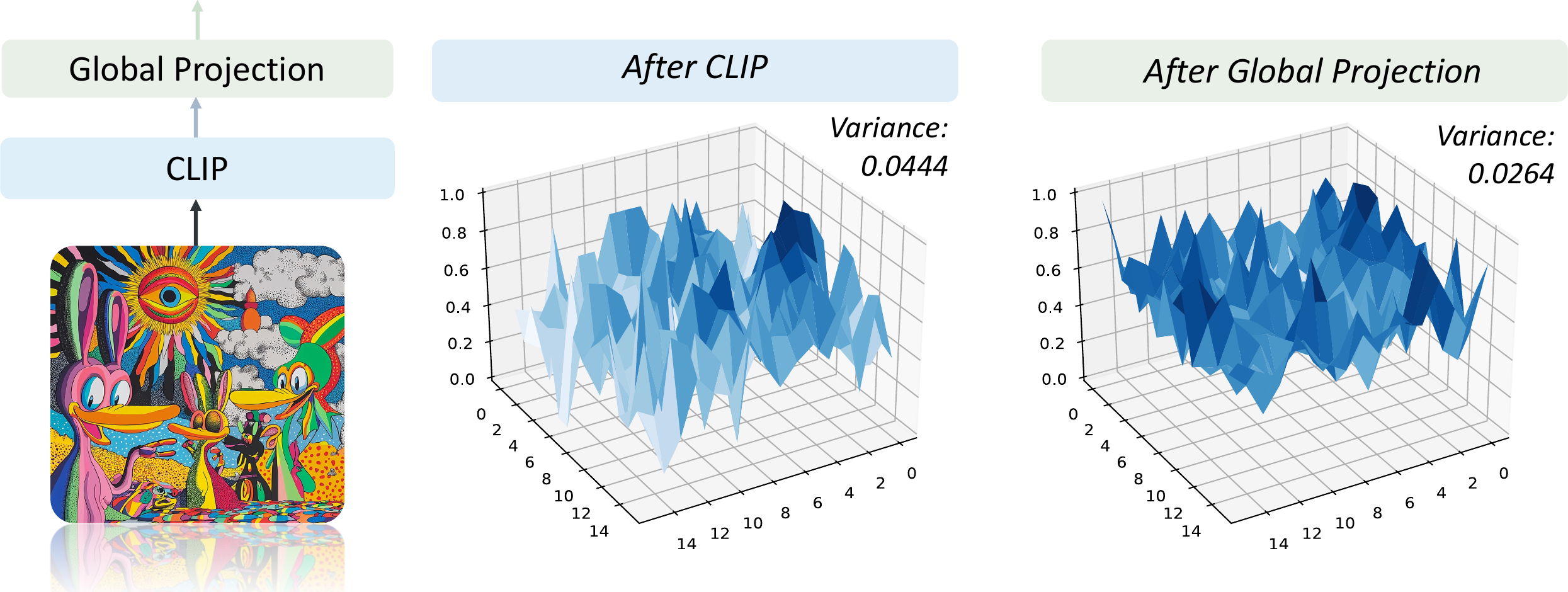}
    \caption{\textbf{Similarity between the extracted global style representations among image patches}. Without our global projection, the CLIP image embedding only attends to specific regions; while after the projection, the attention shows an even distribution.}
    \label{fig:global}
\end{figure}

This projection allows us to preserve the generalization capabilities of pre-trained CLIP while tailoring the output to focus on style-oriented features. The MLP layer serves as a learnable transformation that distills style information from the image embedding. 
As illustrated in Fig.~\ref{fig:global}, we compare the global feature's similarity with patches before and after the projection.
The result shows that, compared to similarities between global feature and patch features before projection showing peak distribution, the after-projection ones are more evenly distributed. This is also supported by the smaller variance after projection. 
It aligns with the target of the global style description, which should represent the whole image. Therefore, we obtain the global feature by:
\begin{equation}
    F_\text{global} = \text{MLP}(F_i) \nonumber.
\end{equation}

\subsection{Combine Local and Global Description}
\label{ssec:local}

\begin{figure}
    \centering
    \includegraphics[width=1\linewidth]{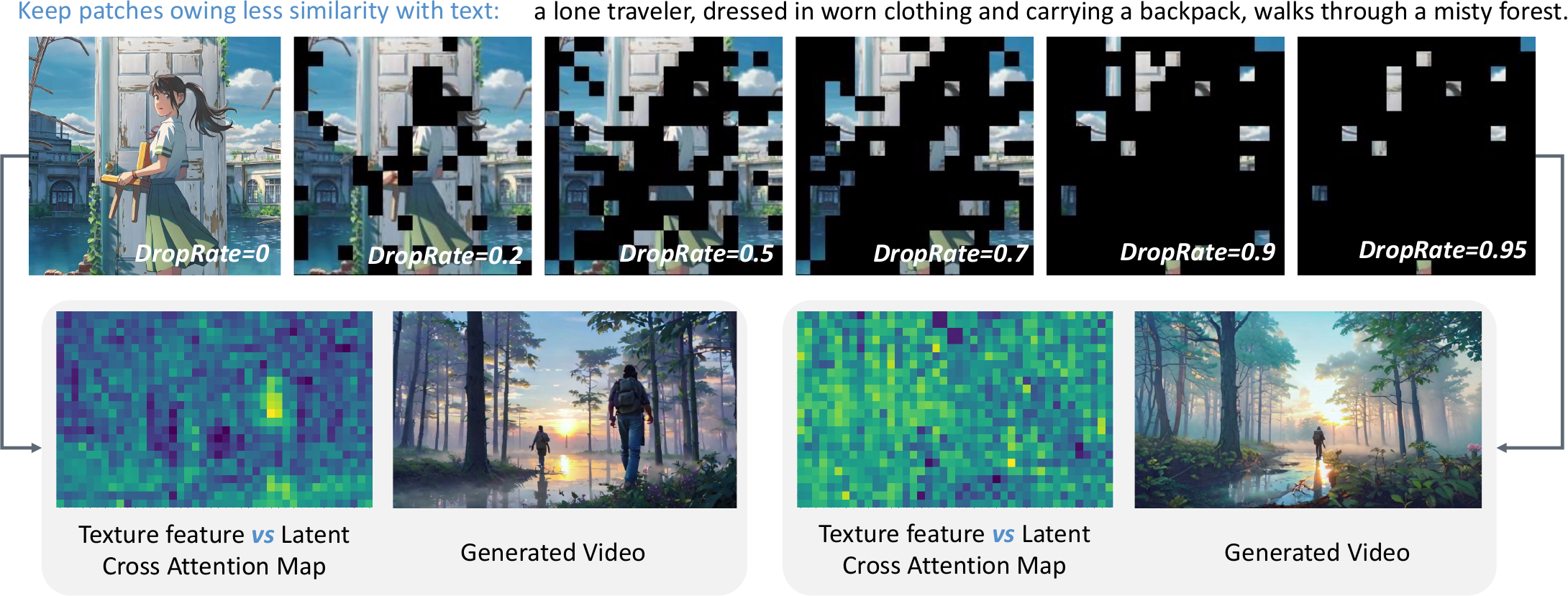}
    \caption{\textbf{The selection of texture feature using similarity with prompt features}. Top: the kept patches under different drop rates, showing that the dropped tokens are mainly on human-related regions (especially when the drop rate is $0.7$). Bottom: the attention map of the cross-attention between texture feature and latent when the drop rate is $0$ and $0.95$, and their generated results.}
    \label{fig:loca}
\end{figure}

However, relying solely on the global description is insufficient for obtaining optimal style representations. As illustrated in Fig.~\ref{fig:texture}, while the global representation can accurately capture the overall style of the reference image, it fails to preserve local textures, such as the distinctive brushstrokes in Van Gogh paintings. To address this limitation, we consider to use the patch features extracted by CLIP:
\begin{equation}
F_p = \text{CLIP}(I).\texttt{patch\_feature}\,, \nonumber
\end{equation}
where $F_p \in \mathbb{R}^{256 \times 1024}$. However, directly preserving all patch features would risk content leakage. Therefore, we propose a selection strategy to choose only a few patch features as the texture feature. To avoid any content leakage, we incorporate the prompt feature, and compare it with image patch features to obtain similarity scores. We further choose the patches sharing less similarity score with the texture feature, which are more likely to carry only texture instead of any content. Specifically, we choose them by
\begin{equation}
F_p' = \text{concat}({F_p^i \mid i \in \text{argsort}(\text{similarity}(F_p, F_{text}))[:k]})\,, \nonumber
\end{equation}
and $k$ is set to $15$ in our method. Following StyleCrafter, we use Q-Former~\cite{blip2} structure to further gather features from the filtered patches. We create $N$ learnable tokens $F_{query} \in \mathbb{R}^{N \times C}$ and then concatenate it with $F_p'$ to perform self-attention~\cite{vaswani2017attention}:
\begin{equation}
F_{{attn}} = \text{self-attention}(\text{concat}(F_{{query}}, F_p'))\,. \nonumber
\end{equation}
Then we take out the first $N$ tokens from $F_{attn}$ as $F_{texture}\in\mathbb{R}^{ N \times C}$ as the texture feature. As shown in Fig.~\ref{fig:loca}, the first row demonstrates the kept patches with varying drop ratios, the patches of the face and body are gradually dropped due to higher similarity with the prompt, which includes the description of a human. Additionally, without selection, directly using all patches will pose interruptions to text alignment. For example, when the drop rate is set to $0$, the texture only attends to the person on the right, serving as content guidance instead of texture guidance.

The texture feature $F_{texture}$ and the global style description $F_{global}$ are concatenated as $F_{style}$, to perform the dual-cross-attention in an adapter manner~\cite{ipadapter}, as 
\begin{equation}
F_{out} = \text{TCA}(F_{in},F_{text})+\text{SCA}(F_{in},F_{style})\,, \nonumber
\end{equation}
where $\text{TCA}$ represents text cross-attention, and $\text{SCA}$ refers to style cross-attention, the $F_{in}$ is the input of cross-attention module, and $F_{out}$ is the output.

\subsection{Motion Adapter for Temporal and Style Quality}
\label{ssec:motion}
While the aforementioned design enables us to inject the style information into model, it would result in temporal flickering and limited range of dynamics. To address these issues, we propose a method to enhance temporal quality with minimal modifications, inspired by the success of Still-Moving \cite{stillmoving}. It demonstrates a smooth transition from customized T2I (Text-to-Image) models to customized T2V (Text-to-Video) models by incorporating a motion adapter trained on still videos. Specifically, for each weight matrix $W \in \{W_Q, W_K, W_V\}$ in the temporal attention block, we train a LoRA~\cite{lora} by applying the following transformation:
\begin{equation}
\widetilde{W} = W + \alpha \cdot A_{t}^{W, \ \text{down}} \cdot A_{t}^{W, \ \text{up}}\,, \nonumber
\end{equation}
where $A_{t}^{W,\ \text{down}}$ and $A_{t}^{W,\ \text{up}}$ are learnable parameters of the motion adapter, trained on still videos with $\alpha = 1$. This formulation offers flexibility in controlling the model's behavior. Setting $\alpha = 0$ leaves the original model unchanged. Setting $\alpha = 1$ generates static videos. Setting $\alpha = -1$ produces the opposite effect, transitioning from stillness to a greater dynamic range. More importantly, since we train the adapter on real-world videos, setting $\alpha = -1$ not only increases the dynamic range but also enhances the stylization by moving further away from the real-world domain, aligning the goal of stylization.

\begin{figure*}[!t]
    \centering
    \includegraphics[width=1\linewidth]{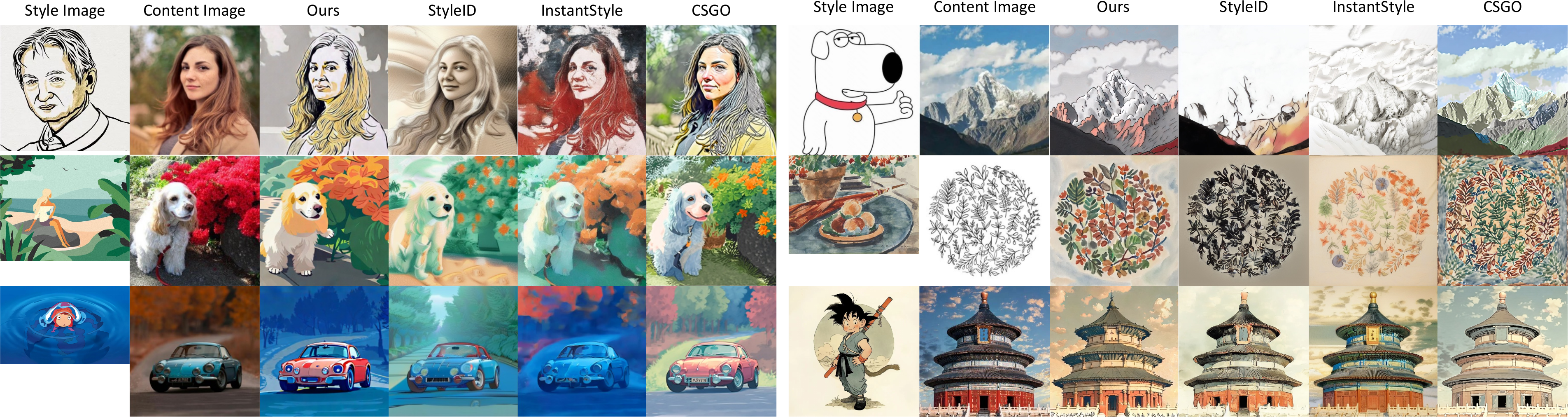}
    \caption{\textbf{Uncurated image style transfer results}. We compare with the recent state-of-the-art methods InstantStyle~\cite{instantstyle}, StyleID~\cite{styleid} and CSGO~\cite{csgo}. Best viewed in Color.}
    \label{fig:image-transfer-comparison}
\end{figure*}

\subsection{Gray Tile ControlNet}
\label{ssec:gray}
To enable both stylized generation and style transfer, we incorporate content guidance into our model. Following CSGO~\cite{csgo} and InstantStylePlus~\cite{instantstyleplus}, we employ a tile ControlNet as the content guidance mechanism. However, we find that the color information in the tile image may interfere with the style transfer process, as shown in Fig.~\ref{fig:content-control}. To address this, we remove the color information from the tile image, converting it into a grayscale image.

The gray tile ControlNet uses $N/2$ vanilla DiT blocks, which inject the content feature into the denoising network at regular intervals. The vanilla DiT block only contains self-attention, temporal attention, text cross-attention, and FFN, and does not include specific designs like the motion adapter and style cross-attention. The output from each vallina DiT block will be added to the corresponding style DiT block as the content guidance.

\section{Experiments}
\label{sec:exp}

\noindent\textbf{Implementation Details}. We develop a DiT-based~\cite{dit} video generation model as our base model, which consists of a 3D casual VAE and several DiT Blocks as the denoising network, as shown in Fig~\ref{fig:structure}. We first train the global style extractor on 10K pairs of style data generated by the model illusion through contrastive learning. Then, we train the motion adapter with still videos for about $300$ iterations with batch size $64$. Next, we start to train the style modulation on image dataset, \textit{i.e.}, Laion-Aesthetics~\cite{schuhmann2022laion} with a batch size of $160$ per GPU for 40K iterations. With the style module ready, we train the gray tile ControlNet with the image dataset, using the above setting for 20K iterations. We train our model on $8$ A800 GPUs, which can be completed within two days. For the classifier-free guidance, we use the decoupled cfg like other methods~\cite{stylecrafter}. In our method, we set text cfg to $12.5$ and style cfg to $6$.

\noindent\textbf{Evaluation Metrics}. For the image style transfer task, we employ the metric CSD score~\cite{csd} used in CSGO~\cite{csgo} to measure style similarity with the reference image, and the series of metrics used in StyleID~\cite{styleid} to validate the style transfer quality. Specifically, ArtFID \cite{artfid} is notable for its strong correlation with human perception, as it considers both style and content fidelity. 
We also adopt the CFSD metric~\cite{styleid} to further validate the content preservation.

For the assessment of video stylization, we employ a two-fold validation. First, we utilize image metrics to perform frame-by-frame validation of image stylization. Second, following StyleCrafter~\cite{stylecrafter}, we evaluate motion quality. Specifically, we adopt motion smoothness metric and dynamic degree metric proposed in VBench~\cite{huang2024vbench} for evaluation, using the AMT~\cite{li2023amt} and RAFT~\cite{teed2020raft} as the base model respectively. For the text-video alignment, we employ UMT score~\cite{umt} and CLIP-Text~\cite{clip} similarity as metrics.

\noindent\textbf{Dataset}.
For video stylized generation, we base our test set on that proposed by StyleCrafter~\cite{stylecrafter}. We expand this set by adding more style images and prompts, obtaining a comprehensive test set comprising $12$ style images and $16$ prompts, yielding a total of $192$ style-prompt pairs.

For image style transfer, we curate a test set consisting of $8$ content images paired with the aforementioned $12$ style images. This leads to $96$ content-style pairs, matching the test set size used in other image style transfer methods~\cite{styleid}.

\begin{table}[!t]
\centering
\footnotesize
\addtolength{\tabcolsep}{-1pt}
\begin{tabular}{@{}lcccc@{}}
\toprule
& \makecell{StyleID~\cite{styleid}\\(CVPR'24)} & \makecell{InstantStyle~\cite{instantstyle}\\(arxiv'24)} & \makecell{CSGO~\cite{csgo}\\(arxiv'24)}  &Ours \\
\midrule
\textbf{CSD-Score} $\uparrow$ & 0.40 & 0.32 & 0.35  &\textbf{0.45} \\
\textbf{ArtFID} $\downarrow$ & 38.57 & 42.48 & 41.42  &\textbf{36.89} \\
FID $\downarrow$ & 23.91 & 24.59 & 25.71  &\textbf{22.11} \\
LPIPS $\downarrow$ & \textbf{0.55} & 0.67 & 0.56  &0.61 \\
CFSD $\downarrow$ & \textbf{1.06} & 1.70 & 5.12  &2.37 \\
\bottomrule
\end{tabular}
\caption{\textbf{The quantitative results of image style transfer}. ArtFID considers both style resemble and content preservation. CSD-score represents the similarity of style. The last two reflect content preservation. The first two metrics in bold are the most representative metrics for this evaluation.}
\label{tab:style_comparison}
\end{table}

\begin{figure*}[!t]
    \centering
    \includegraphics[width=1\linewidth]{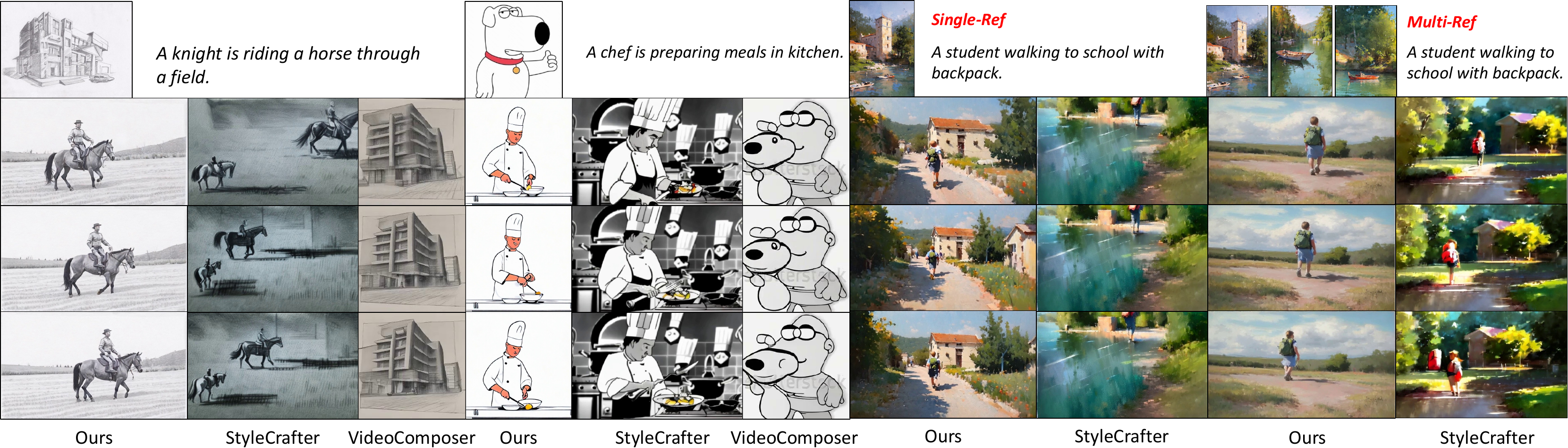}
    \caption{\textbf{Qualitative comparison of single-reference and multi-reference style-guided T2V generation}. We compare with StyleCrafter~\cite{stylecrafter} and VideoComposer~\cite{wang2024videocomposer}. Best viewed in color.}
    \label{fig:video-generation}
\end{figure*}

\subsection{Image Style Transfer}
We consider the image style transfer as the most intuitive evaluation method for style learning, with minimal dependence on base model generation abilities or temporal factors. Therefore, we also compare our method with image stylization methods by regarding the image as a one-frame video. We choose two training-free SOTA methods StyleID~\cite{styleid} and InstantStyle~\cite{instantstyle}, and a training-based SOTA method CSGO as our competitors. In Table~\ref{tab:style_comparison}, following CSGO~\cite{csgo} and StyleID~\cite{styleid}, we demonstrate the CSD scores~\cite{csd} and the content preservation metrics of the proposed method compared to recent advanced methods. Our method significantly outperforms others in the first three metrics, indicating accurate style learning from reference images. While slightly underperforming in content alignment metrics, we argue that effective style transfer requires balancing style fidelity and content retention. For example, as shown in Fig.~\ref{fig:image-transfer-comparison} (top row), the Noble style causes some loss of details due to its simple line style, which is an expected transformation to fit the style. However, we argue that effective style transfer is not solely judged by content preservation, but rather on achieving an optimal balance between style fidelity and content retention. The ArtFID metric, which aligns well with human preference, shows our method's significant advantage. Figure~\ref{fig:image-transfer-comparison} presents uncurated test samples, demonstrating our method's ability to capture reference style accurately while maintaining high content preservation.

\begin{table}[!t]
\centering
\footnotesize
\renewcommand{\arraystretch}{1}
\addtolength{\tabcolsep}{-4pt}
\begin{tabular}{@{}lccc@{}}

\toprule
 &  \makecell{VideoComposer~\cite{wang2024videocomposer}\\ \footnotesize (NeurIPS'24)} &\makecell{StyleCrafter~\cite{stylecrafter}\\ \footnotesize(SIGGRAPH Asia'24)} & Ours\\
\midrule
CLIP-Text $\uparrow$ &  0.057 &0.294 & \textbf{0.305} \\
UMT-Score $\uparrow$ &  -2.268 &1.994 & \textbf{2.329} \\
CSD-Score $\uparrow$ &  \textbf{0.680} &0.448 & 0.463 \\
VisualQuality $\uparrow$ &  2.159 &2.140 & \textbf{2.370} \\
DynamicQuality $\uparrow$ &  2.284 &2.306 & \textbf{2.496} \\
MotionSmooth $\uparrow$ &  0.975 &0.973 & \textbf{0.994} \\
\bottomrule
\end{tabular}
\caption{\textbf{Comparison of stylized video generation results}. We compare our method with VideoComposer~\cite{wang2024videocomposer} and StyleCrafter~\cite{stylecrafter}. Our method demonstrates higher style resemblance and stronger text alignment.}
\label{tab:stylized-video-comparison}
\end{table}

\begin{figure}[!t]
    \centering
    \includegraphics[width=1\linewidth]{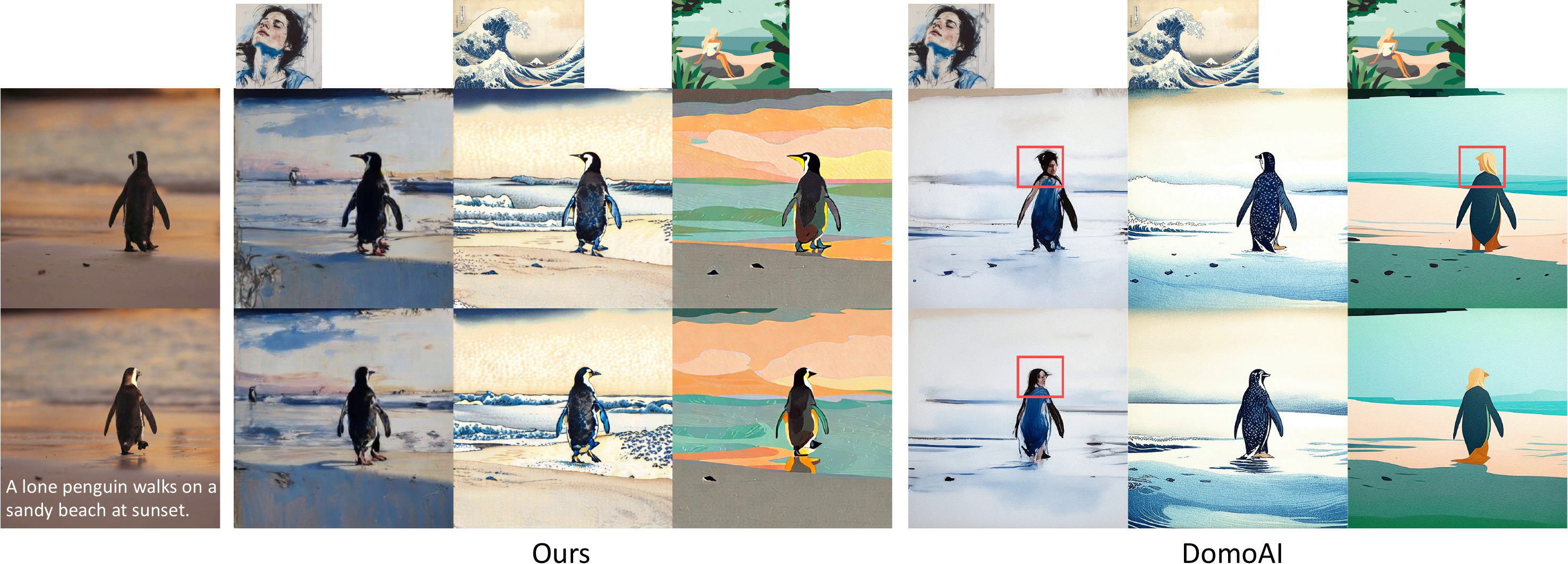}
    \caption{\textbf{Video style transfer results compared with DomoAI}. Their results disrupt semantics, shown in red bounding box.}
    \label{fig:domo}
\end{figure}

\subsection{Stylized Video Generation}

For the stylized video generation task, we use the aforementioned test set which contains $192$ style-prompt pairs to generate videos. We compare our method with previous state-of-the-art methods StyleCrafter~\cite{stylecrafter} and VideoComposer~\cite{wang2024videocomposer}. As shown in Table~\ref{tab:stylized-video-comparison}, our method outperforms these two methods in five metrics, which demonstrates our superiority in alignment between text and video ($0.305$ CLIP-Text similarity and $2.329$ UMT-Score), enhanced visual and dynamic quality, and smoother motion. Our CSD score falls behind VideoComposer. The reason lies in that it directly copies the content in the reference image, therefore exhibiting a higher style score. Instead, our method implements style injection on the basis of text alignment, and achieves both high T2V alignment and high style consistency with the reference image.

The visualization results are shown in Fig.~\ref{fig:video-generation}. Other methods either fail to accurately capture the style in the reference image or suffer from poor text alignment. For example, the generation results of VideoComposer almost show no correspondence with the given prompt, which aligns with the negative UMT score in Table~\ref{tab:stylized-video-comparison}. Although StyleCrafter demonstrates style similarity to some extent, it learns only the superficial style representations like color but not the complete style descriptions. We also compare the generation results with single/multiple reference images, both generating high-quality stylized videos.

\renewcommand{\thefootnote}{\arabic{footnote}}
\subsection{Video Style Transfer}
Here we compare our method with an online commercial application DomoAI\footnote{https://www.domoai.app/}. As shown in Fig.~\ref{fig:domo}, it can achieve appealing stylization results, but they may interrupt the semantics within the video.

\subsection{Ablation Studies}

\subsubsection{Style Representation Extraction}
To validate whether our specific designs in the style extraction module really matter, we conduct an ablation study to show their effect. The results are reported in Table~\ref{tab:ablation_global}. Comparing B1 with B2, the use of Global Project (GP) can effectively prevent content leakage caused by directly using the CLIP image embedding, with an obvious improvement in the UMT score ($2.337$ vs $0.892$). Additionally, directly using all image patch features as texture features will bring a similar problem, reflected by the poor $0.771$ UMT score in B3. However, selecting only a few of them can help alleviate the problem; \textit{i.e.}, randomly discarding most tokens can enhance text alignment. Furthermore, comparing B4 with B5, if we select the tokens by considering their similarity with prompts instead of a random selection, we can obtain higher text alignment while maintaining style similarity. Variant B6 demonstrates that the integration of global and local styles can further enhance style resemblance, achieving a CSD score of $0.463$.

\begin{table}[!t]
\centering
\small
\begin{tabular}{@{}l|cc|cc|cc@{}}
\toprule
 & \multicolumn{2}{c|}{Global}& \multicolumn{2}{c|}{Texture}& UMT&CSD\\

&w/o GP&w/ GP&w/o S&w/ S& Score & Score\\ \midrule

   B1&\checkmark&&&& 0.892& 0.561\\
   B2&&\checkmark&&& 2.337& 0.443\\
   B3&&&\checkmark&& 0.771& 0.534\\
 B4& & & & \texttt{random}& 2.129&0.454\\
   B5&&&&\checkmark& 2.331& 0.452\\
   B6&&\checkmark&&\checkmark& 2.329& 0.463\\
\bottomrule
\end{tabular}
\caption{\textbf{The ablation study of the style extraction module design}. GP means the global projection after CLIP image embedding, S refers to the selection using text features. $\texttt{random}$ represents selecting the same amount of tokens randomly.}
\label{tab:ablation_global}
\end{table}

\subsubsection{Motion Adapter}
To explore the effect of motion adapter on both dynamics and style, we conduct an ablation study with different $\alpha$ values ranging from $0$ to $-1$. As shown in Table~\ref{table:ma}, when the negative scale of motion adapter increases ($0 \rightarrow -1$), the CSD score representing style similarity also increases. It verifies that, due to the training on real-world images, the negative ratio can generate results away from the real-world domain, leading to a more stylistic video. Also, the dynamic degree will increase as the scale increases. However, it damages the text alignment (UMTScore) and the motion smooth when the scale exceeds $0.3$. Therefore, we choose $-0.3$ as our setting, which owns the best visual quality and also well achieves the dynamic degree and style similarity.

\begin{table}[!t]
\small
\centering
\begin{tabular}{@{}l|ccccc@{}}
\toprule
MotionAdapterScale &   -1&-0.5&\textbf{-0.3} & -0.1 & 0 \\ \midrule
CSDScore $\uparrow$ &   0.465&0.465&0.463& 0.446&0.443\\
DynamicDegree$\uparrow$&   20.559&9.320&6.579 & 1.576 &1.371  \\
UMTScore$\uparrow$&   2.211&2.193&2.329 & 2.235 &2.272 \\
MotionSmooth$\uparrow$ &   0.990&0.992&0.994 & 0.994 &0.994 \\
VisualQuality$\uparrow$ &  2.259&2.263&2.370 &2.279 &2.278 \\
\bottomrule
\end{tabular}
\caption{\textbf{The effect of motion adapter on generation results}. We choose $-0.3$ as the suitable scale.}
\label{table:ma}
\end{table}

\begin{figure}[!t]
    \centering
    \includegraphics[width=1\linewidth]{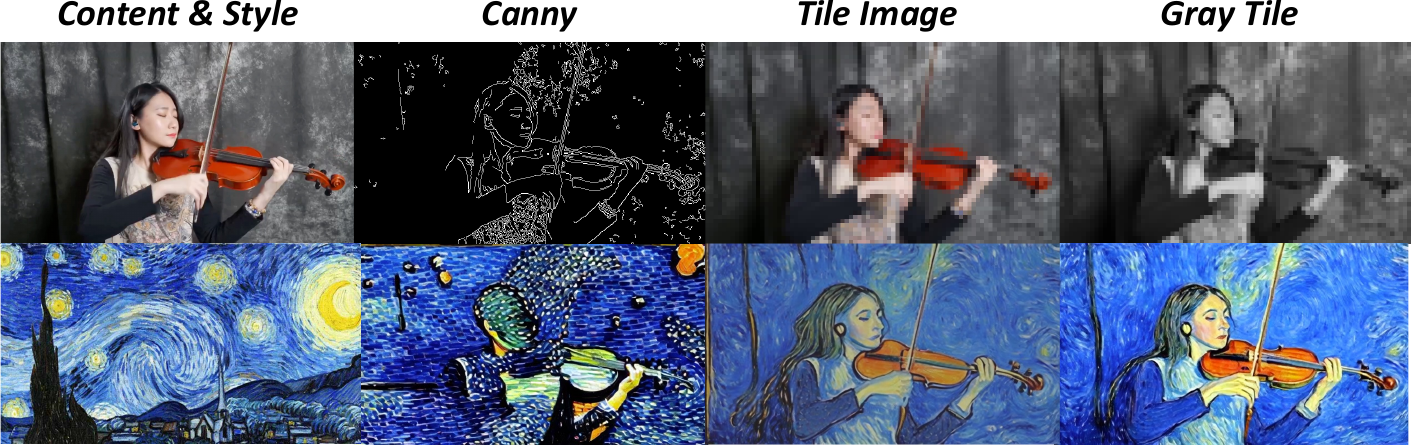}
    \caption{\textbf{Ablation of different conditions of ControlNet in our method}. The gray tile achieves the best performance.}
    \label{fig:content-control}
\end{figure}

\subsubsection{Content Control}
We compare different conditions for content control during generation. We compare gray tile guidance with Canny and RGB tile images. As shown in Fig.~\ref{fig:content-control}, the Canny method provides too much detailed information but less layout guidance. In contrast, the RGB tile image can provide a layout hint, leading to more precise content control. However, the color information within the guidance can interrupt the style injection, resulting in darker outputs, as shown in the third column. To alleviate this, we use the gray tile image as the condition. The results verify the improvements.

\section{Conclusion}

In this paper, we address the challenges faced by existing stylization methods, particularly in sub-optimal style extraction and the lack of video translation. To tackle these issues, we propose a novel approach that leverages both global and local style representations to achieve an ideal style descriptor. Our method involves selecting local patches with minimal content similarity to capture texture details and using a contrastive learning strategy to train a global style extractor with paired data generated through model illusion. To enhance video quality, we incorporate a motion adapter, which improves motion quality and style extent during inference. Additionally, we implement a gray tile ControlNet for more precise content guidance in video translation tasks. Beyond the implementation, our method significantly outperforms other methods in both text alignment and style resemblance. 

\noindent\textbf{Limitation}. Current stylization methods typically rely on reference style images. However, video stylization includes more than just graphic style—it also involves dynamic elements like particle effects and motion characteristics. In future research, we aim to explore methods for extracting and transferring dynamic styles from reference videos.

{
    \small
    \bibliographystyle{ieeenat_fullname}
    \bibliography{main}

\begin{thebibliography}{54}
\providecommand{\natexlab}[1]{#1}
\providecommand{\url}[1]{\texttt{#1}}
\expandafter\ifx\csname urlstyle\endcsname\relax
  \providecommand{\doi}[1]{doi: #1}\else
  \providecommand{\doi}{doi: \begingroup \urlstyle{rm}\Url}\fi

\bibitem[Brooks et~al.(2024)Brooks, Peebles, Holmes, DePue, Guo, Jing, Schnurr, Taylor, Luhman, Luhman, et~al.]{sora}
Tim Brooks, Bill Peebles, Connor Holmes, Will DePue, Yufei Guo, Li Jing, David Schnurr, Joe Taylor, Troy Luhman, Eric Luhman, et~al.
\newblock Video generation models as world simulators, 2024.

\bibitem[Cai et~al.(2024)Cai, Liu, Zhang, and Shi]{caiclap}
Yichao Cai, Yuhang Liu, Zhen Zhang, and Javen~Qinfeng Shi.
\newblock Clap: Isolating content from style through contrastive learning with augmented prompts.
\newblock \emph{European conference on computer vision}, 2024.

\bibitem[Canny(1986)]{canny}
John Canny.
\newblock A computational approach to edge detection.
\newblock \emph{IEEE Transactions on pattern analysis and machine intelligence}, pages 679--698, 1986.

\bibitem[Chefer et~al.(2024)Chefer, Zada, Paiss, Ephrat, Tov, Rubinstein, Wolf, Dekel, Michaeli, and Mosseri]{stillmoving}
Hila Chefer, Shiran Zada, Roni Paiss, Ariel Ephrat, Omer Tov, Michael Rubinstein, Lior Wolf, Tali Dekel, Tomer Michaeli, and Inbar Mosseri.
\newblock Still-moving: Customized video generation without customized video data.
\newblock \emph{arXiv preprint arXiv:2407.08674}, 2024.

\bibitem[Chen et~al.(2017)Chen, Liao, Yuan, Yu, and Hua]{chen2017coherent}
Dongdong Chen, Jing Liao, Lu Yuan, Nenghai Yu, and Gang Hua.
\newblock Coherent online video style transfer.
\newblock In \emph{Proceedings of the IEEE International Conference on Computer Vision}, pages 1105--1114, 2017.

\bibitem[Chen et~al.(2024{\natexlab{a}})Chen, Tennent, and Hsu]{artadapter}
Dar-Yen Chen, Hamish Tennent, and Ching-Wen Hsu.
\newblock Artadapter: Text-to-image style transfer using multi-level style encoder and explicit adaptation.
\newblock In \emph{Proceedings of the IEEE/CVF Conference on Computer Vision and Pattern Recognition}, pages 8619--8628, 2024{\natexlab{a}}.

\bibitem[Chen et~al.(2023)Chen, Xia, He, Zhang, Cun, Yang, Xing, Liu, Chen, Wang, Weng, and Shan]{videocrafter1}
Haoxin Chen, Menghan Xia, Yingqing He, Yong Zhang, Xiaodong Cun, Shaoshu Yang, Jinbo Xing, Yaofang Liu, Qifeng Chen, Xintao Wang, Chao Weng, and Ying Shan.
\newblock Videocrafter1: Open diffusion models for high-quality video generation, 2023.

\bibitem[Chen et~al.(2024{\natexlab{b}})Chen, Laina, and Vedaldi]{layout}
Minghao Chen, Iro Laina, and Andrea Vedaldi.
\newblock Training-free layout control with cross-attention guidance.
\newblock In \emph{Proceedings of the IEEE/CVF Winter Conference on Applications of Computer Vision}, pages 5343--5353, 2024{\natexlab{b}}.

\bibitem[Chen et~al.(2024{\natexlab{c}})Chen, Huang, Liu, Shen, Zhao, and Zhao]{anydoor}
Xi Chen, Lianghua Huang, Yu Liu, Yujun Shen, Deli Zhao, and Hengshuang Zhao.
\newblock Anydoor: Zero-shot object-level image customization.
\newblock In \emph{Proceedings of the IEEE/CVF Conference on Computer Vision and Pattern Recognition}, pages 6593--6602, 2024{\natexlab{c}}.

\bibitem[Chung et~al.(2024)Chung, Hyun, and Heo]{styleid}
Jiwoo Chung, Sangeek Hyun, and Jae-Pil Heo.
\newblock Style injection in diffusion: A training-free approach for adapting large-scale diffusion models for style transfer.
\newblock In \emph{Proceedings of the IEEE/CVF Conference on Computer Vision and Pattern Recognition}, pages 8795--8805, 2024.

\bibitem[Deng et~al.(2021)Deng, Tang, Dong, Huang, Ma, and Xu]{deng2021arbitrary}
Yingying Deng, Fan Tang, Weiming Dong, Haibin Huang, Chongyang Ma, and Changsheng Xu.
\newblock Arbitrary video style transfer via multi-channel correlation.
\newblock In \emph{Proceedings of the AAAI Conference on Artificial Intelligence}, pages 1210--1217, 2021.

\bibitem[Frenkel et~al.(2024)Frenkel, Vinker, Shamir, and Cohen-Or]{blora}
Yarden Frenkel, Yael Vinker, Ariel Shamir, and Daniel Cohen-Or.
\newblock Implicit style-content separation using b-lora.
\newblock \emph{arXiv preprint arXiv:2403.14572}, 2024.

\bibitem[Gal et~al.(2022)Gal, Alaluf, Atzmon, Patashnik, Bermano, Chechik, and Cohen-Or]{ti}
Rinon Gal, Yuval Alaluf, Yuval Atzmon, Or Patashnik, Amit~H Bermano, Gal Chechik, and Daniel Cohen-Or.
\newblock An image is worth one word: Personalizing text-to-image generation using textual inversion.
\newblock \emph{arXiv preprint arXiv:2208.01618}, 2022.

\bibitem[Gao et~al.(2020)Gao, Li, Yin, and Yang]{gao2020fast}
Wei Gao, Yijun Li, Yihang Yin, and Ming-Hsuan Yang.
\newblock Fast video multi-style transfer.
\newblock In \emph{Proceedings of the IEEE/CVF winter conference on applications of computer vision}, pages 3222--3230, 2020.

\bibitem[Geng et~al.(2024)Geng, Park, and Owens]{geng2024illusion}
Daniel Geng, Inbum Park, and Andrew Owens.
\newblock Visual anagrams: Generating multi-view optical illusions with diffusion models.
\newblock In \emph{Proceedings of the IEEE/CVF Conference on Computer Vision and Pattern Recognition}, pages 24154--24163, 2024.

\bibitem[Guo et~al.(2024)Guo, Yang, Rao, Liang, Wang, Qiao, Agrawala, Lin, and Dai]{animatediff}
Yuwei Guo, Ceyuan Yang, Anyi Rao, Zhengyang Liang, Yaohui Wang, Yu Qiao, Maneesh Agrawala, Dahua Lin, and Bo Dai.
\newblock Animatediff: Animate your personalized text-to-image diffusion models without specific tuning.
\newblock \emph{International Conference on Learning Representations}, 2024.

\bibitem[Ho et~al.(2020)Ho, Jain, and Abbeel]{ho2020denoising}
Jonathan Ho, Ajay Jain, and Pieter Abbeel.
\newblock Denoising diffusion probabilistic models.
\newblock \emph{Advances in neural information processing systems}, 33:\penalty0 6840--6851, 2020.

\bibitem[Hong et~al.(2022)Hong, Ding, Zheng, Liu, and Tang]{cogvideo}
Wenyi Hong, Ming Ding, Wendi Zheng, Xinghan Liu, and Jie Tang.
\newblock Cogvideo: Large-scale pretraining for text-to-video generation via transformers.
\newblock \emph{arXiv preprint arXiv:2205.15868}, 2022.

\bibitem[Hu et~al.(2021)Hu, Shen, Wallis, Allen-Zhu, Li, Wang, Wang, and Chen]{lora}
Edward~J Hu, Yelong Shen, Phillip Wallis, Zeyuan Allen-Zhu, Yuanzhi Li, Shean Wang, Lu Wang, and Weizhu Chen.
\newblock Lora: Low-rank adaptation of large language models.
\newblock \emph{arXiv preprint arXiv:2106.09685}, 2021.

\bibitem[Huang et~al.(2017)Huang, Wang, Luo, Ma, Jiang, Zhu, Li, and Liu]{huang2017real}
Haozhi Huang, Hao Wang, Wenhan Luo, Lin Ma, Wenhao Jiang, Xiaolong Zhu, Zhifeng Li, and Wei Liu.
\newblock Real-time neural style transfer for videos.
\newblock In \emph{IEEE Conference on Computer Vision and Pattern Recognition (CVPR)}, pages 7044--7052, 2017.

\bibitem[Huang et~al.(2024)Huang, He, Yu, Zhang, Si, Jiang, Zhang, Wu, Jin, Chanpaisit, et~al.]{huang2024vbench}
Ziqi Huang, Yinan He, Jiashuo Yu, Fan Zhang, Chenyang Si, Yuming Jiang, Yuanhan Zhang, Tianxing Wu, Qingyang Jin, Nattapol Chanpaisit, et~al.
\newblock Vbench: Comprehensive benchmark suite for video generative models.
\newblock In \emph{Proceedings of the IEEE/CVF Conference on Computer Vision and Pattern Recognition}, pages 21807--21818, 2024.

\bibitem[Ku et~al.(2024)Ku, Wei, Ren, Yang, and Chen]{anyv2v}
Max Ku, Cong Wei, Weiming Ren, Huan Yang, and Wenhu Chen.
\newblock Anyv2v: A plug-and-play framework for any video-to-video editing tasks.
\newblock \emph{arXiv preprint arXiv:2403.14468}, 2024.

\bibitem[Li et~al.(2023{\natexlab{a}})Li, Li, Savarese, and Hoi]{blip2}
Junnan Li, Dongxu Li, Silvio Savarese, and Steven Hoi.
\newblock Blip-2: Bootstrapping language-image pre-training with frozen image encoders and large language models.
\newblock In \emph{International conference on machine learning}, pages 19730--19742. PMLR, 2023{\natexlab{a}}.

\bibitem[Li et~al.(2024{\natexlab{a}})Li, Fang, Zou, Gong, Zheng, Wang, Chen, and Yang]{styletokenizer}
Wen Li, Muyuan Fang, Cheng Zou, Biao Gong, Ruobing Zheng, Meng Wang, Jingdong Chen, and Ming Yang.
\newblock Styletokenizer: Defining image style by a single instance for controlling diffusion models.
\newblock \emph{arXiv preprint arXiv:2409.02543}, 2024{\natexlab{a}}.

\bibitem[Li et~al.(2023{\natexlab{b}})Li, Zhu, Han, Hou, Guo, and Cheng]{li2023amt}
Zhen Li, Zuo-Liang Zhu, Ling-Hao Han, Qibin Hou, Chun-Le Guo, and Ming-Ming Cheng.
\newblock Amt: All-pairs multi-field transforms for efficient frame interpolation.
\newblock In \emph{Proceedings of the IEEE/CVF Conference on Computer Vision and Pattern Recognition}, pages 9801--9810, 2023{\natexlab{b}}.

\bibitem[Li et~al.(2024{\natexlab{b}})Li, Cao, Wang, Qi, Cheng, and Shan]{li2024photomaker}
Zhen Li, Mingdeng Cao, Xintao Wang, Zhongang Qi, Ming-Ming Cheng, and Ying Shan.
\newblock Photomaker: Customizing realistic human photos via stacked id embedding.
\newblock In \emph{Proceedings of the IEEE/CVF Conference on Computer Vision and Pattern Recognition}, pages 8640--8650, 2024{\natexlab{b}}.

\bibitem[Lipman et~al.(2022)Lipman, Chen, Ben-Hamu, Nickel, and Le]{lipman2022flow}
Yaron Lipman, Ricky~TQ Chen, Heli Ben-Hamu, Maximilian Nickel, and Matt Le.
\newblock Flow matching for generative modeling.
\newblock \emph{arXiv preprint arXiv:2210.02747}, 2022.

\bibitem[Liu et~al.(2023)Liu, Xia, Zhang, Chen, Xing, Wang, Wang, Yang, and Shan]{stylecrafter}
Gongye Liu, Menghan Xia, Yong Zhang, Haoxin Chen, Jinbo Xing, Yibo Wang, Xintao Wang, Yujiu Yang, and Ying Shan.
\newblock Stylecrafter: Enhancing stylized text-to-video generation with style adapter.
\newblock \emph{arXiv preprint arXiv:2312.00330}, 2023.

\bibitem[Liu et~al.(2022)Liu, Li, Wu, Chen, Shan, and Qie]{umt}
Ye Liu, Siyuan Li, Yang Wu, Chang-Wen Chen, Ying Shan, and Xiaohu Qie.
\newblock Umt: Unified multi-modal transformers for joint video moment retrieval and highlight detection.
\newblock In \emph{Proceedings of the IEEE/CVF Conference on Computer Vision and Pattern Recognition}, pages 3042--3051, 2022.

\bibitem[Ma et~al.(2024{\natexlab{a}})Ma, Goldstein, Albergo, Boffi, Vanden-Eijnden, and Xie]{ma2024sit}
Nanye Ma, Mark Goldstein, Michael~S Albergo, Nicholas~M Boffi, Eric Vanden-Eijnden, and Saining Xie.
\newblock Sit: Exploring flow and diffusion-based generative models with scalable interpolant transformers.
\newblock \emph{arXiv preprint arXiv:2401.08740}, 2024{\natexlab{a}}.

\bibitem[Ma et~al.(2024{\natexlab{b}})Ma, Wang, Jia, Chen, Liu, Li, Chen, and Qiao]{latte}
Xin Ma, Yaohui Wang, Gengyun Jia, Xinyuan Chen, Ziwei Liu, Yuan-Fang Li, Cunjian Chen, and Yu Qiao.
\newblock Latte: Latent diffusion transformer for video generation.
\newblock \emph{arXiv preprint arXiv:2401.03048}, 2024{\natexlab{b}}.

\bibitem[Mou et~al.(2024)Mou, Wang, Xie, Wu, Zhang, Qi, and Shan]{t2iadapter}
Chong Mou, Xintao Wang, Liangbin Xie, Yanze Wu, Jian Zhang, Zhongang Qi, and Ying Shan.
\newblock T2i-adapter: Learning adapters to dig out more controllable ability for text-to-image diffusion models.
\newblock In \emph{Proceedings of the AAAI Conference on Artificial Intelligence}, pages 4296--4304, 2024.

\bibitem[Peebles and Xie(2023)]{dit}
William Peebles and Saining Xie.
\newblock Scalable diffusion models with transformers.
\newblock In \emph{Proceedings of the IEEE/CVF International Conference on Computer Vision}, pages 4195--4205, 2023.

\bibitem[Qi et~al.(2024)Qi, Fang, Wu, Xie, Liu, Chen, He, and Zhang]{qi2024deadiff}
Tianhao Qi, Shancheng Fang, Yanze Wu, Hongtao Xie, Jiawei Liu, Lang Chen, Qian He, and Yongdong Zhang.
\newblock Deadiff: An efficient stylization diffusion model with disentangled representations.
\newblock In \emph{Proceedings of the IEEE/CVF Conference on Computer Vision and Pattern Recognition}, pages 8693--8702, 2024.

\bibitem[Radford et~al.(2021)Radford, Kim, Hallacy, Ramesh, Goh, Agarwal, Sastry, Askell, Mishkin, Clark, et~al.]{clip}
Alec Radford, Jong~Wook Kim, Chris Hallacy, Aditya Ramesh, Gabriel Goh, Sandhini Agarwal, Girish Sastry, Amanda Askell, Pamela Mishkin, Jack Clark, et~al.
\newblock Learning transferable visual models from natural language supervision.
\newblock In \emph{International conference on machine learning}, pages 8748--8763. PMLR, 2021.

\bibitem[Rombach et~al.(2022)Rombach, Blattmann, Lorenz, Esser, and Ommer]{stablediffusion}
Robin Rombach, Andreas Blattmann, Dominik Lorenz, Patrick Esser, and Bj{\"o}rn Ommer.
\newblock High-resolution image synthesis with latent diffusion models.
\newblock In \emph{Proceedings of the IEEE/CVF conference on computer vision and pattern recognition}, pages 10684--10695, 2022.

\bibitem[Ruiz et~al.(2023)Ruiz, Li, Jampani, Pritch, Rubinstein, and Aberman]{dreambooth}
Nataniel Ruiz, Yuanzhen Li, Varun Jampani, Yael Pritch, Michael Rubinstein, and Kfir Aberman.
\newblock Dreambooth: Fine tuning text-to-image diffusion models for subject-driven generation.
\newblock In \emph{Proceedings of the IEEE/CVF conference on computer vision and pattern recognition}, pages 22500--22510, 2023.

\bibitem[Schuhmann et~al.(2022)Schuhmann, Beaumont, Vencu, Gordon, Wightman, Cherti, Coombes, Katta, Mullis, Wortsman, et~al.]{schuhmann2022laion}
Christoph Schuhmann, Romain Beaumont, Richard Vencu, Cade Gordon, Ross Wightman, Mehdi Cherti, Theo Coombes, Aarush Katta, Clayton Mullis, Mitchell Wortsman, et~al.
\newblock Laion-5b: An open large-scale dataset for training next generation image-text models.
\newblock \emph{Advances in Neural Information Processing Systems}, 35:\penalty0 25278--25294, 2022.

\bibitem[Sohn et~al.(2023)Sohn, Ruiz, Lee, Chin, Blok, Chang, Barber, Jiang, Entis, Li, et~al.]{styledrop}
Kihyuk Sohn, Nataniel Ruiz, Kimin Lee, Daniel~Castro Chin, Irina Blok, Huiwen Chang, Jarred Barber, Lu Jiang, Glenn Entis, Yuanzhen Li, et~al.
\newblock Styledrop: Text-to-image generation in any style.
\newblock \emph{arXiv preprint arXiv:2306.00983}, 2023.

\bibitem[Somepalli et~al.(2024)Somepalli, Gupta, Gupta, Palta, Goldblum, Geiping, Shrivastava, and Goldstein]{csd}
Gowthami Somepalli, Anubhav Gupta, Kamal Gupta, Shramay Palta, Micah Goldblum, Jonas Geiping, Abhinav Shrivastava, and Tom Goldstein.
\newblock Measuring style similarity in diffusion models.
\newblock \emph{arXiv preprint arXiv:2404.01292}, 2024.

\bibitem[Song et~al.(2020)Song, Sohl-Dickstein, Kingma, Kumar, Ermon, and Poole]{song2020score}
Yang Song, Jascha Sohl-Dickstein, Diederik~P Kingma, Abhishek Kumar, Stefano Ermon, and Ben Poole.
\newblock Score-based generative modeling through stochastic differential equations.
\newblock \emph{arXiv preprint arXiv:2011.13456}, 2020.

\bibitem[Teed and Deng(2020)]{teed2020raft}
Zachary Teed and Jia Deng.
\newblock Raft: Recurrent all-pairs field transforms for optical flow.
\newblock In \emph{Computer Vision--ECCV 2020: 16th European Conference, Glasgow, UK, August 23--28, 2020, Proceedings, Part II 16}, pages 402--419. Springer, 2020.

\bibitem[Vaswani(2017)]{vaswani2017attention}
A Vaswani.
\newblock Attention is all you need.
\newblock \emph{Advances in Neural Information Processing Systems}, 2017.

\bibitem[Wang et~al.(2024{\natexlab{a}})Wang, Spinelli, Wang, Bai, Qin, and Chen]{instantstyle}
Haofan Wang, Matteo Spinelli, Qixun Wang, Xu Bai, Zekui Qin, and Anthony Chen.
\newblock Instantstyle: Free lunch towards style-preserving in text-to-image generation.
\newblock \emph{arXiv preprint arXiv:2404.02733}, 2024{\natexlab{a}}.

\bibitem[Wang et~al.(2024{\natexlab{b}})Wang, Xing, Huang, Ai, Wang, and Bai]{instantstyleplus}
Haofan Wang, Peng Xing, Renyuan Huang, Hao Ai, Qixun Wang, and Xu Bai.
\newblock Instantstyle-plus: Style transfer with content-preserving in text-to-image generation.
\newblock \emph{arXiv preprint arXiv:2407.00788}, 2024{\natexlab{b}}.

\bibitem[Wang et~al.(2024{\natexlab{c}})Wang, Yuan, Zhang, Chen, Wang, Zhang, Shen, Zhao, and Zhou]{wang2024videocomposer}
Xiang Wang, Hangjie Yuan, Shiwei Zhang, Dayou Chen, Jiuniu Wang, Yingya Zhang, Yujun Shen, Deli Zhao, and Jingren Zhou.
\newblock Videocomposer: Compositional video synthesis with motion controllability.
\newblock \emph{Advances in Neural Information Processing Systems}, 36, 2024{\natexlab{c}}.

\bibitem[Wang et~al.(2023)Wang, Wang, Xie, Qi, Shan, Wang, and Luo]{styleadapter}
Zhouxia Wang, Xintao Wang, Liangbin Xie, Zhongang Qi, Ying Shan, Wenping Wang, and Ping Luo.
\newblock Styleadapter: A single-pass lora-free model for stylized image generation.
\newblock \emph{arXiv preprint arXiv:2309.01770}, 2023.

\bibitem[Wright and Ommer(2022)]{artfid}
Matthias Wright and Bj{\"o}rn Ommer.
\newblock Artfid: Quantitative evaluation of neural style transfer.
\newblock In \emph{DAGM German Conference on Pattern Recognition}, pages 560--576. Springer, 2022.

\bibitem[Xie et~al.(2023)Xie, Li, Huang, Liu, Zhang, Zheng, and Shou]{boxdiff}
Jinheng Xie, Yuexiang Li, Yawen Huang, Haozhe Liu, Wentian Zhang, Yefeng Zheng, and Mike~Zheng Shou.
\newblock Boxdiff: Text-to-image synthesis with training-free box-constrained diffusion.
\newblock In \emph{Proceedings of the IEEE/CVF International Conference on Computer Vision}, pages 7452--7461, 2023.

\bibitem[Xing et~al.(2024{\natexlab{a}})Xing, Liu, Xia, Zhang, Wang, Shan, and Wong]{tooncrafter}
Jinbo Xing, Hanyuan Liu, Menghan Xia, Yong Zhang, Xintao Wang, Ying Shan, and Tien-Tsin Wong.
\newblock Tooncrafter: Generative cartoon interpolation.
\newblock \emph{arXiv preprint arXiv:2405.17933}, 2024{\natexlab{a}}.

\bibitem[Xing et~al.(2024{\natexlab{b}})Xing, Wang, Sun, Wang, Bai, Ai, Huang, and Li]{csgo}
Peng Xing, Haofan Wang, Yanpeng Sun, Qixun Wang, Xu Bai, Hao Ai, Renyuan Huang, and Zechao Li.
\newblock Csgo: Content-style composition in text-to-image generation.
\newblock \emph{arXiv preprint arXiv:2408.16766}, 2024{\natexlab{b}}.

\bibitem[Ye et~al.(2023)Ye, Zhang, Liu, Han, and Yang]{ipadapter}
Hu Ye, Jun Zhang, Sibo Liu, Xiao Han, and Wei Yang.
\newblock Ip-adapter: Text compatible image prompt adapter for text-to-image diffusion models.
\newblock \emph{arXiv preprint arXiv:2308.06721}, 2023.

\bibitem[Zhang et~al.(2023{\natexlab{a}})Zhang, Rao, and Agrawala]{controlnet}
Lvmin Zhang, Anyi Rao, and Maneesh Agrawala.
\newblock Adding conditional control to text-to-image diffusion models.
\newblock In \emph{Proceedings of the IEEE/CVF International Conference on Computer Vision}, pages 3836--3847, 2023{\natexlab{a}}.

\bibitem[Zhang et~al.(2023{\natexlab{b}})Zhang, Huang, Tang, Huang, Ma, Dong, and Xu]{inst}
Yuxin Zhang, Nisha Huang, Fan Tang, Haibin Huang, Chongyang Ma, Weiming Dong, and Changsheng Xu.
\newblock Inversion-based style transfer with diffusion models.
\newblock In \emph{Proceedings of the IEEE/CVF conference on computer vision and pattern recognition}, pages 10146--10156, 2023{\natexlab{b}}.

\end{thebibliography}
}

\clearpage
\setcounter{page}{1}
\maketitlesupplementary

\section{Overview}
\label{sec:overview}

This Supplementary Material is organized into four sections, providing additional details and results to complement the main paper:

\begin{itemize}
\item \textbf{Section~\ref{sec:imple}:} Provides comprehensive implementation details, including the structure of the base model and the illusion dataset construction.
\item \textbf{Section~\ref{sec:image-style-transfer}:} Illustrates complete results of image style transfer.
\item \textbf{Section~\ref{sec:stylized-video-generation}:} Showcases additional results for stylized video generation.
\item \textbf{Section~\ref{sec:video-style-transfer}:} Offers a quantitative comparison of video style transfer results with DomoAI.
\end{itemize}

\section{Implementation Details}
\label{sec:imple}
\subsection{Base Model Structure}

\begin{figure}[htbp]
    \centering
    \includegraphics[width=1\linewidth]{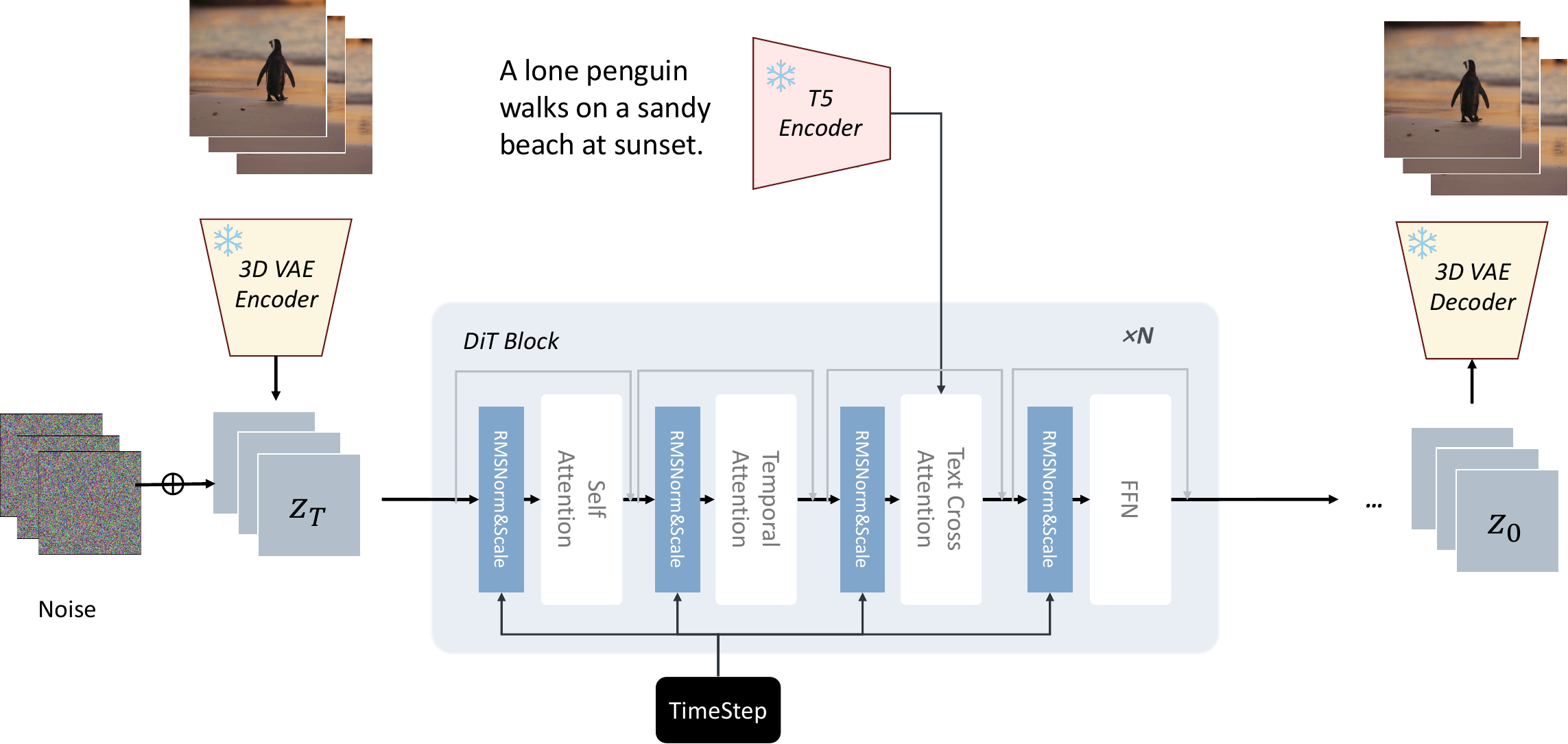}
    \caption{The structure of our base model.}
    \label{fig:structure}
\end{figure}

Our model is a DiT-based structure, which consists of a 3D Variational AutoEncoder to convert the video to latent space. Then, the latent feature will pass several DiT blocks~\cite{dit}. As shown in Fig.~\ref{fig:structure}, each DiT block contains 2D Self-Attention, 3D Self-Attention, Cross-Attention and FFN module. The timestep is embedded as scale and apply RMSNorm to the spatio-temporal tokens before each module.

\subsection{Model Illusion Dataset}
First, we will make a detailed analysis on illusion process.

\noindent \textbf{Model Illusion Process.} During the generation process using an off-the-shelf T2I model, we can use two different prompts to generate paired image data. To be specific, for a noisy image, we can start a parallel process.

\begin{figure}[!t]
    \centering
    \includegraphics[width=1\linewidth]{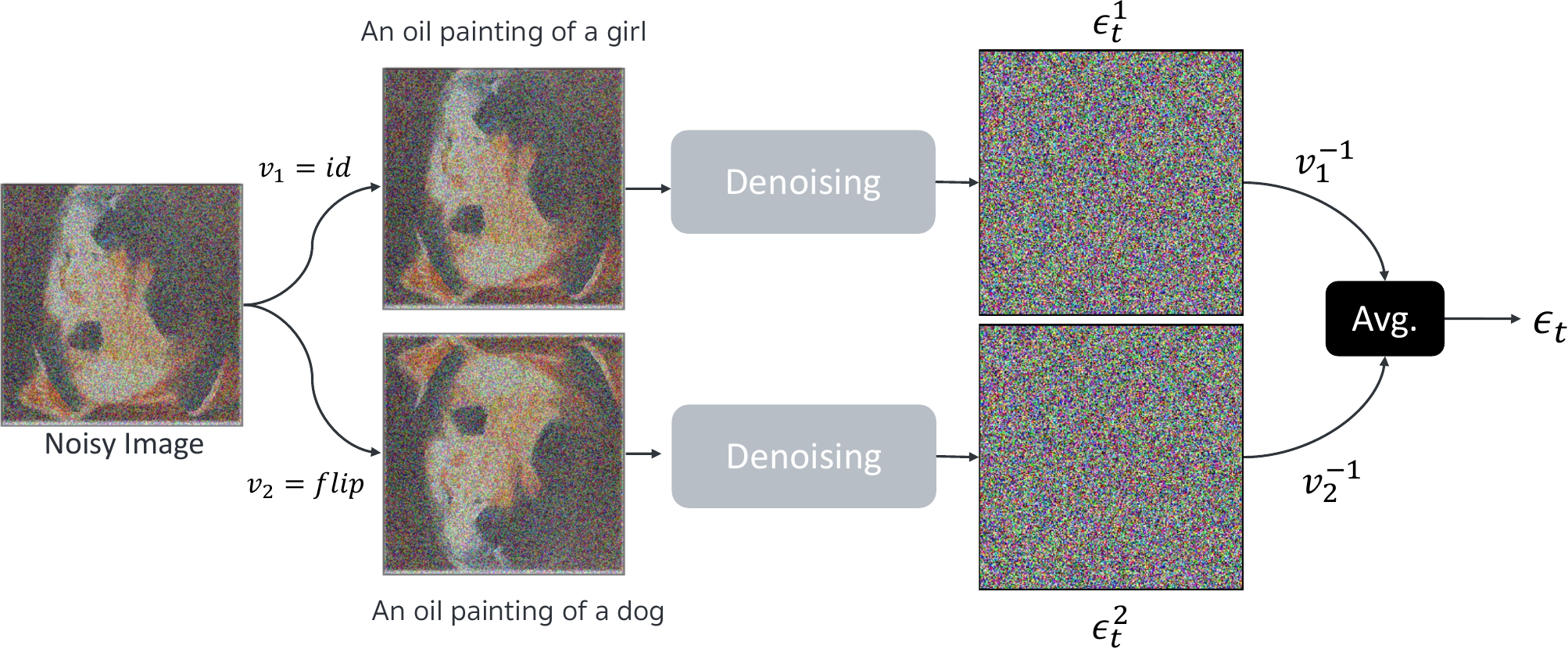}
    \caption{The model illusion process during T2I generation.}
    \label{fig:illusion-process}
\end{figure}

As shown in the Fig.~\ref{fig:illusion-process}, we conduct two transformations on it respectively, in the figure, we use the original image and the vertical flip, which are defined by $v_1$ and $v_2$. Then, we use different text prompts to guide the dual-denoising, and we can obtain $\epsilon_t^1$ and $\epsilon_t^2$. Then, we use $v_1^{-1}$ and $v_2^{-1}$ to turn $\epsilon_t^1$ and $\epsilon_t^2$ to the original view, i.e., the view of noisy image. For $\epsilon_t^1$, since $v_1$ is just itself, so $v_1^{-1}$ will not change $\epsilon_t^1$. For $\epsilon_t^2$, because $v_2$ is vertical clip, so $v_2^{-1}$ will perform a reversed vertical clip. Then the reversed noise in original view will be added and averaged, to obtain the final $\epsilon_t$.

During our generation, we set $v_1 = id$ and $v_2 = jigsaw$, the $jigsaw$ means to divide the image into irregular puzzle pieces.

\noindent \textbf{Prompt.} For each paired images, we use a pair of prompts to generate them. The prompts are in form of \textit{a [style] of [object]}. The style and object are from our style list which contains $65$ style descriptions collected from the Internet, and the object list consists of $100$ common objects. Here we demonstrate $15$ samples of each list.

\noindent \textbf{Training.} The training loop runs for 100 epochs with a batch size of 8. Two losses are used to supervise this process: triplet loss, which increases the distance between groups, and MSE loss, which reduces the distance within groups. The learning rate is set to 1e-4.

\begin{table}[htbp]
    \centering
    \begin{tabular}{c|c}
        \toprule
         object&  style\\
         \midrule
        a dog& oil painting\\
        a rabbit& black and white film\\
        a waterfall& cyberpunk picture\\
        a duck& watercolor painting\\
        a teddy bear& vintage photograph\\
        a tudor portrait& 3D render\\
        a skull& pencil sketch\\
        houseplants& pop art depiction\\
        flowers& surrealist painting\\
        a landscape& pixel art representation\\
        a boy& comic book illustration\\
        a girl& graffiti street art\\
        an ape& neon-lit cityscape\\
        a parrot& Baroque art piece\\
        a panda& steampunk design\\
        \bottomrule
    \end{tabular}
    \caption{Samples from the object list and style list we use for illusion dataset.}
    \label{tab:list_sample}
\end{table}

\section{Image Style Transfer}
\label{sec:image-style-transfer}

Since our method can also be used as an image stylization method, so we compare our method with other image stylization methods, including StyleID~\cite{styleid}, InstantStyle~\cite{instantstyle} and CSGO~\cite{csgo}. We use the default setting in these methods.

\begin{figure*}[!t]
    \centering
    \includegraphics[width=1\linewidth]{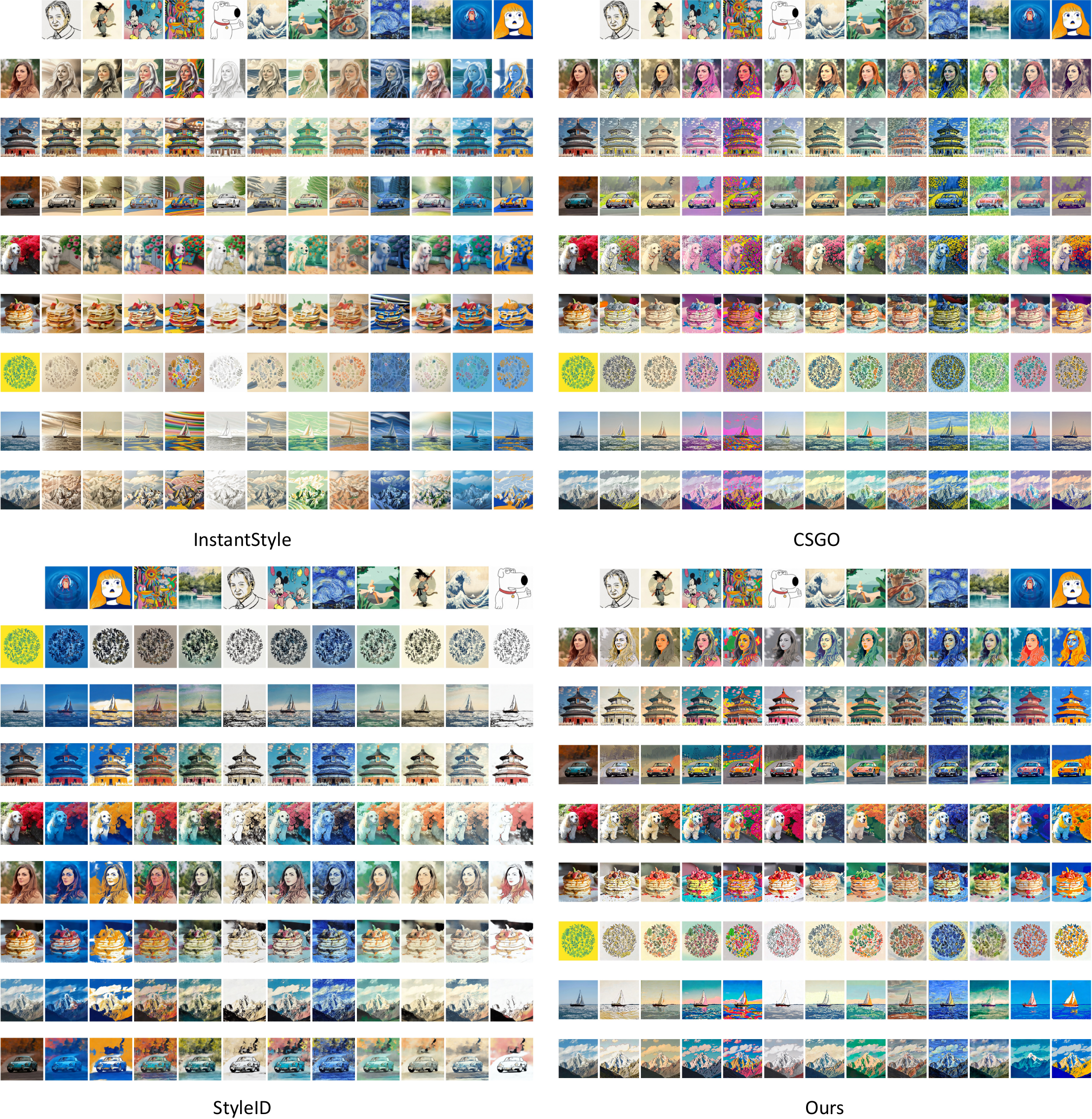}
    \caption{The image style transfer results generated by four different methods.}
    \label{fig:complete-image}
\end{figure*}

Here we illustrate all image style transfer results generated by these three methods with our method in Fig.~\ref{fig:complete-image}. It is obvious that our method show higher robustness to different styles. For example, StyleID~\cite{styleid} show great ability in transferring the colors, to keep color consistency with reference image. However, it cannot transfer the sematic features of the style.

\begin{table}[!t]
    \centering
    \begin{tabular}{|p{0.45\textwidth}|} %
        \hline
        A beautiful woman with long, wavy, white hair stands outdoors.\\
        \hline
        A stunning view of Temple of Heaven in Beijing, showcasing its intricate architecture and against a backdrop of a partly cloudy sky.\\
        \hline
        A classic car is parked on a winding road with trees in the background.\\
        \hline
        A fluffy dog sits on a sidewalk next to a vibrant bush of flowers, with rocks in the background.\\
        \hline
        A stack of pancakes topped with strawberries, nuts, and a drizzle of caramel sauce, served with a side of whipped cream on a plate.\\
        \hline
        A circular arrangement of various hand-drawn botanical elements, including leaves, branches, and flowers, set against a pure background.\\
        \hline
        A sailboat glides across sparkling waters under a clear sky, with distant land visible on the horizon.\\
        \hline
        A majestic snow-capped mountain peak rises against a backdrop of a clear sky dotted with fluffy clouds. The rugged terrain and sharp ridges of the mountain are highlighted by the sunlight, creating a stunning contrast with the shadowed valleys below.\\
        \hline
    \end{tabular}
    \caption{The image prompt for image style transfer. Corresponding pictures can refer to Fig.~\ref{fig:complete-image}.}
    \label{tab:prompt_image}
\end{table}

\section{Stylized Video Generation}
\label{sec:stylized-video-generation}
\begin{table}[!t]
    \centering
    \begin{tabular}{|p{0.45\textwidth}|} %
        \hline
        A little girl is reading a book in the beautiful garden.\\
        \hline
        A lighthouse is beaming across choppy waters.\\
        \hline
        A bear is catching fish in a river.\\
        \hline
        A bouquet of fresh flowers sways gently in the vase with the breeze.\\
        \hline
        A rowboat is bobbing on the raging lake.\\
        \hline
        A street performer playing the guitar.\\
        \hline
        A chef is preparing meals in kitchen.\\
        \hline
        A student is walking to school with backpack.\\
        \hline
        A campfire surrounded by tents.\\
        \hline
        A hot air balloon floating in the sky.\\
        \hline
        A knight is riding a horse through a field.\\
        \hline
        A wolf is walking stealthily through the forest.\\
        \hline
        A river is flowing gently under a bridge.\\
        \hline
        A lone traveler, dressed in worn clothing and carrying a backpack, walks through a misty forest at sunset. The trees surrounding him are tall and dense, with leaves that are a mix of green and golden hues, reflecting the warm colors of the setting sun. The mist creates a mystical atmosphere, with the air filled with the sweet scent of blooming flowers. The traveler's footsteps are quiet on the damp earth, and the only sounds are the distant chirping of birds and the rustling of leaves. \\
        \hline
        A group of teddy bears, are holding hands and walking along the street on a rainy day. The bears are dressed in matching raincoats and hats, and they are all smiling and laughing as they stroll along the sidewalk. The rain is coming down gently, and the bears are splashing in the puddles and playing in the rain. The background is a blurred image of the city, with tall buildings and streetlights visible in the distance.\\
        \hline
        A plump rabbit dressed in a vibrant purple robe with intricate golden trimmings walks through a fantastical landscape, with rolling hills, towering trees, and sparkling waterfalls in the background. The landscape is filled with lush greenery, colorful flowers, and towering mushrooms, giving it a whimsical and dreamlike quality. The rabbit's robe is adorned with golden embroidery, and it carries a staff in its hand, giving it a sense of authority and wisdom. The video captures the rabbit's gentle and peaceful movements as it explores this fantastical world, with a soft focus and warm lighting that adds to the sense of wonder and enchantment.\\
        \hline
    \end{tabular}
    \caption{The video prompt for stylized video generation. }
    \label{tab:prompt_video}
\end{table}

\begin{figure*}[!t]
    \centering
    \includegraphics[width=1\linewidth]{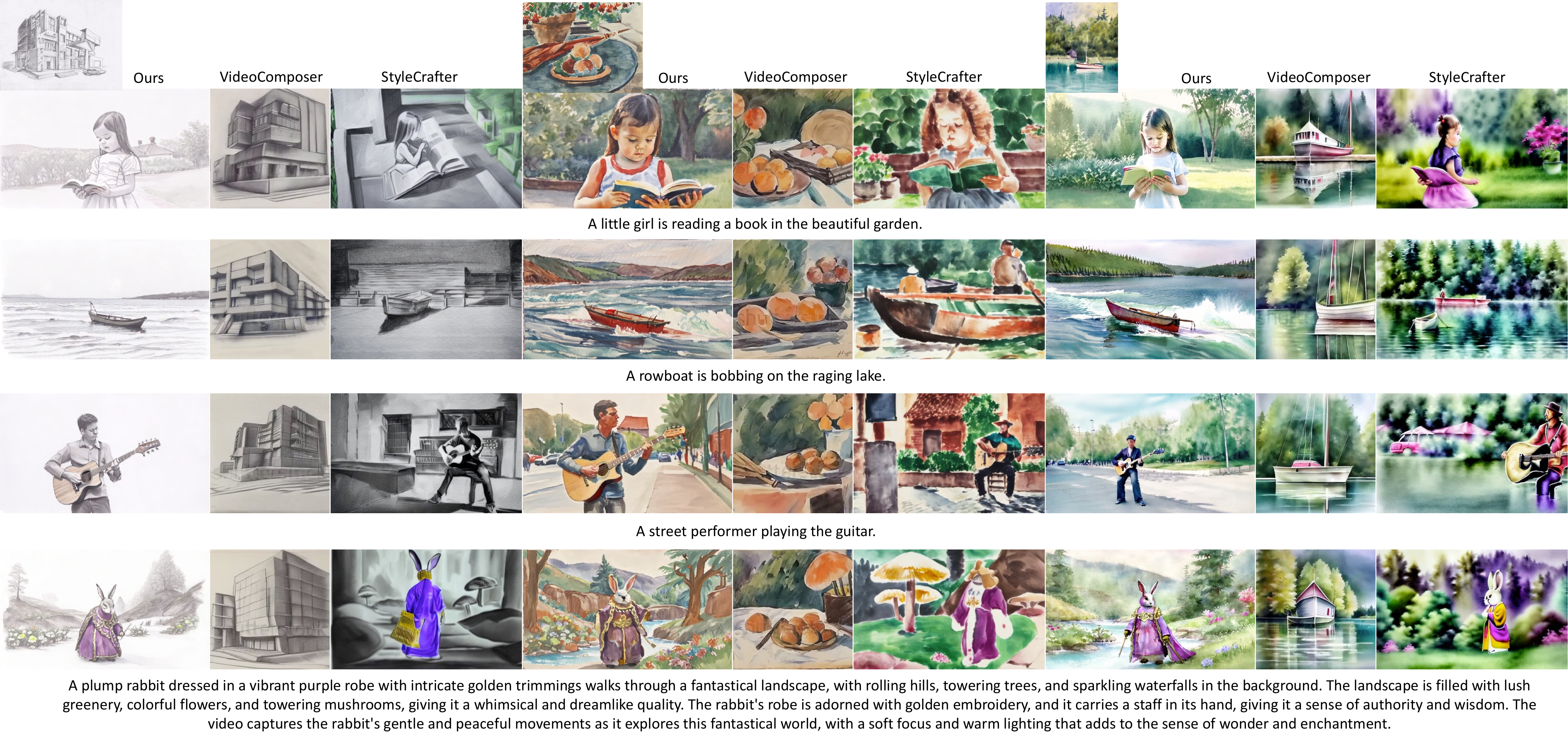}
    \caption{\textbf{More stylized video generation results.} We compare our method with VideoComposer~\cite{wang2024videocomposer} and StyleCrafter~\cite{stylecrafter}.}
    \label{fig:more-stylized-generation}
\end{figure*}

More comparison results are shown in Fig.~\ref{fig:more-stylized-generation}. The compared methods are VideoComposer~\cite{wang2024videocomposer} and StyleCrafter~\cite{stylecrafter}. More videos can be viewed in \url{https://style-master.github.io/}.

\section{Video Style Transfer}

Here we conduct a comparison with DomoAI\footnote{https://www.domoai.app/} and the combination of InstantStyle~\cite{instantstyle} and AnyV2V~\cite{anyv2v}. InstantStyle is used to transfer the style of the first frame, then, AnyV2V will use edited first frame and the video to transfer the whole video. We use the default setting in DomoAI website.

We collect $4$ videos from the Internet, for the comparison with InstantStyle\&AnyV2V, we crop and resize the video to $512\times512$. We use the $16$ style images to test the style transfer ability.
The results are shown in Table~\ref{table:video-style-transfer}, one can see our method has better style transfer ability and better text-alignment score. Also, our visual quality is better.

\label{sec:video-style-transfer}

\begin{table}[!t]
\centering
\begin{tabular}{lccc}
\toprule
 & \textbf{DomoAI} & \textbf{\makecell{InstantStyle\\\&AnyV2V}}  &\textbf{Ours}   \\ \midrule
 CSD-Score$\uparrow$& 0.421&0.313&0.434  \\
 CLIP-Text$\uparrow$& 0.290&0.312&0.319  \\
UMTScore$\uparrow$& 2.349 & 2.984 &3.311  \\
VisualQuality$\uparrow$& 2.331 & 2.158 &2.400  \\
DynamicDegree$\uparrow$& 0.005 & 0.233 &0.082  \\
MotionSmooth$\uparrow$& 0.995 & 0.971 &0.989  \\
\bottomrule
\end{tabular}
\caption{\textbf{Quantitative Comparison with video style transfer methods}, including DomoAI, InstantStyle~\cite{instantstyle} with AnyV2V~\cite{anyv2v}.}
\label{table:video-style-transfer}
\end{table}

\end{document}